\documentclass[]{fairmeta}

\graphicspath{{figures/}}
\usepackage{amsmath}
\usepackage{amssymb}
\usepackage{xspace}
\usepackage{algorithm}
\usepackage{algorithmic}
\usepackage{colortbl}
\usepackage{tabularx}
\usepackage[shortlabels,inline]{enumitem}

\makeatletter
\def\NAT@sort{\@ne}
\def\NAT@cmprs{\@ne}
\makeatother

\newcommand{\method}{\textnormal{\textsc{ViewDreamer}}\xspace}
\newcommand{\benchname}{RoboView-Bench\xspace}
\newcommand{\parahead}[1]{\par\noindent\textbf{#1}\enspace}

\definecolor{resultgreen}{RGB}{0,128,70}
\definecolor{resultred}{RGB}{190,45,45}
\definecolor{oursrow}{gray}{0.92}
\newcommand{\oursrow}{\rowcolor{oursrow}}
\newcommand{\resultdelta}[3]{\begin{tabular}[t]{@{}c@{}}#2\\[-1.5pt]{\scriptsize\textcolor{#1}{(#3)}}\end{tabular}}
\newcommand{\improve}[2]{\resultdelta{resultgreen}{#1}{#2}}
\newcommand{\degrade}[2]{\resultdelta{resultred}{#1}{#2}}
\newcommand{\same}[2]{\resultdelta{gray}{#1}{#2}}

\newcommand{\tablebf}[1]{{\bfseries #1}}
\AtBeginEnvironment{tabular}{\let\textbf\tablebf}
\AtBeginEnvironment{tabular*}{\let\textbf\tablebf}
\AtBeginEnvironment{tabularx}{\let\textbf\tablebf}

\makeatletter
\@namedef{vd@r@1}{1.02588}
\@namedef{vd@r@2}{1.00186}
\@namedef{vd@r@3}{1.01698}
\@namedef{vd@r@4}{1.00766}
\@namedef{vd@r@5}{1.00919}
\@namedef{vd@r@6}{1.00491}
\@namedef{vd@r@7}{1.02177}
\@namedef{vd@r@8}{1.01304}
\@namedef{vd@r@9}{1.05153}
\@namedef{vd@r@10}{1.01930}
\@namedef{vd@r@11}{1.00179}
\@namedef{vd@r@12}{1.05153}
\@namedef{vd@r@13}{1.05153}
\@namedef{vd@r@14}{1.01930}
\@namedef{vd@r@15}{1.00179}
\@namedef{vd@r@16}{1.05153}
\@namedef{vd@r@17}{1.03837}
\@namedef{vd@r@18}{0.99913}
\@namedef{vd@r@19}{1.00485}
\@namedef{vd@r@20}{1.00738}
\@namedef{vd@r@21}{1.01051}
\@namedef{vd@r@22}{0.96495}
\@namedef{vd@r@23}{0.98982}
\@namedef{vd@r@24}{1.04069}
\@namedef{vd@r@25}{0.99768}
\@namedef{vd@r@26}{1.03012}
\@namedef{vd@r@27}{1.02041}
\@namedef{vd@r@28}{1.04325}
\@namedef{vd@r@29}{1.01955}
\@namedef{vd@r@30}{1.04325}
\@namedef{vd@r@31}{1.01955}
\@namedef{vd@r@32}{0.98514}
\@namedef{vd@r@33}{0.99151}
\@namedef{vd@r@34}{0.99361}
\@namedef{vd@r@35}{1.04325}
\@namedef{vd@r@36}{1.02106}
\@namedef{vd@r@37}{1.02163}
\@namedef{vd@r@38}{0.98915}
\@namedef{vd@r@39}{0.98968}
\@namedef{vd@r@40}{0.99134}
\@namedef{vd@r@41}{0.98915}
\@namedef{vd@r@42}{1.01258}
\@namedef{vd@r@43}{1.01418}
\@namedef{vd@r@44}{0.98310}
\@namedef{vd@r@45}{1.00560}
\@namedef{vd@r@46}{1.01256}
\@namedef{vd@r@47}{1.00876}
\@namedef{vd@r@48}{0.99579}
\@namedef{vd@r@49}{1.03163}
\@namedef{vd@r@50}{1.01845}
\@namedef{vd@r@51}{0.99052}
\@namedef{vd@r@52}{1.01861}
\@namedef{vd@r@53}{0.99516}
\@namedef{vd@r@54}{0.99647}
\@namedef{vd@r@55}{1.00851}
\@namedef{vd@r@56}{1.00723}
\@namedef{vd@r@57}{0.99973}
\@namedef{vd@r@58}{0.99768}
\@namedef{vd@r@59}{0.99670}
\@namedef{vd@r@60}{1.00928}
\@namedef{vd@r@61}{0.99874}
\@namedef{vd@r@62}{1.01946}
\@namedef{vd@r@63}{1.01048}
\@namedef{vd@r@64}{1.01064}
\@namedef{vd@r@65}{1.03033}
\@namedef{vd@r@66}{1.05322}
\@namedef{vd@r@67}{1.01345}
\@namedef{vd@r@68}{1.00917}
\@namedef{vd@r@69}{0.98374}
\@namedef{vd@r@70}{1.01460}
\@namedef{vd@r@71}{1.00851}
\@namedef{vd@r@72}{0.99999}
\@namedef{vd@r@73}{1.03334}
\@namedef{vd@r@74}{1.02415}
\@namedef{vd@r@75}{0.98236}
\@namedef{vd@r@76}{1.04490}
\@namedef{vd@r@77}{0.99719}
\@namedef{vd@r@78}{0.99915}
\@namedef{vd@r@79}{0.99882}
\@namedef{vd@r@80}{1.01473}
\@namedef{vd@r@81}{1.01683}
\@namedef{vd@r@82}{0.99560}
\@namedef{vd@r@83}{1.02084}
\@namedef{vd@r@84}{1.04069}
\@namedef{vd@r@85}{1.01257}
\@namedef{vd@r@86}{1.00643}
\@namedef{vd@r@87}{1.02891}
\@namedef{vd@r@88}{1.02650}
\@namedef{vd@r@89}{1.03206}
\@namedef{vd@r@90}{1.01276}
\@namedef{vd@r@91}{1.07203}
\@namedef{vd@r@92}{1.02496}
\@namedef{vd@r@93}{1.01035}
\@namedef{vd@r@94}{1.01360}
\@namedef{vd@r@95}{1.01558}
\@namedef{vd@r@96}{1.01473}
\@namedef{vd@r@97}{1.01276}
\@namedef{vd@r@98}{0.99835}
\@namedef{vd@r@99}{1.02650}
\@namedef{vd@r@100}{1.01209}
\@namedef{vd@r@101}{1.00459}
\@namedef{vd@r@102}{1.01345}
\@namedef{vd@r@103}{0.99443}
\@namedef{vd@r@104}{1.01662}

\newcount\vd@n
\NewCommandCopy\vd@origbf\textbf
\DeclareRobustCommand{\textbf}[1]{%
  \global\advance\vd@n\@ne
  \ifcsname vd@r@\the\vd@n\endcsname
    \edef\vd@stretch{\noexpand\addfontfeatures{FakeStretch=\csname vd@r@\the\vd@n\endcsname}}%
    \vd@origbf{\vd@stretch#1}%
  \else\vd@origbf{#1}\fi}
\makeatother

\title{ViewDreamer: A Robot World Model beyond Fixed Camera Setups}

\author[1,2,*]{Ye Mao}
\author[1]{Jinhui Xiong}
\author[1]{Egor Larionov}
\author[1]{Shiyang Lu}
\author[2]{Krystian Mikolajczyk}
\author[1]{Wei Ye}

\affiliation[1]{Meta AI}
\affiliation[2]{Imperial College London}

\contribution[*]{Work done at Meta}

\abstract{\looseness=-1 Video generation models are increasingly important for robot learning, yet most generate each camera view separately or tile all views in a fixed layout. Robot demonstrations, however, are captured by synchronized cameras whose number, placement, and order vary across embodiments. We present \method, a single world model that handles variable sets of cameras by disentangling per-view generation from cross-view interaction. Each view follows the generation path of a pretrained single-view model, and views interact only through shared slot tokens. We stabilize multi-view training with a curriculum that gradually increases the number of views. Cross-View Slot Agreement uses visual foundation models to group geometrically and semantically similar regions across views into shared slots. On \benchname, which covers one to four cameras, \method achieves the best generation quality among the models evaluated and improves cross-view consistency over independent per-view generation. These gains carry over to our world-action model (WAM) and action-conditioned world model (AC-WM). Our WAM outperforms the FastWAM-Joint baseline on LIBERO-Plus, RoboTwin 2.0, and RoboDojo, and on a real Franka robot. Both variants improve out-of-domain generalization to various camera settings and scenes.\par
}

\date{November 2026}
\correspondence{\email{ye.mao21@imperial.ac.uk}, \email{weiye@meta.com}}
\metadata[Project page]{\url{https://matchlab-imperial.github.io/ViewDreamer/}}

\begin{document}

\maketitle

\begin{figure}[!htb]
    \centering
    \includegraphics[width=\linewidth]{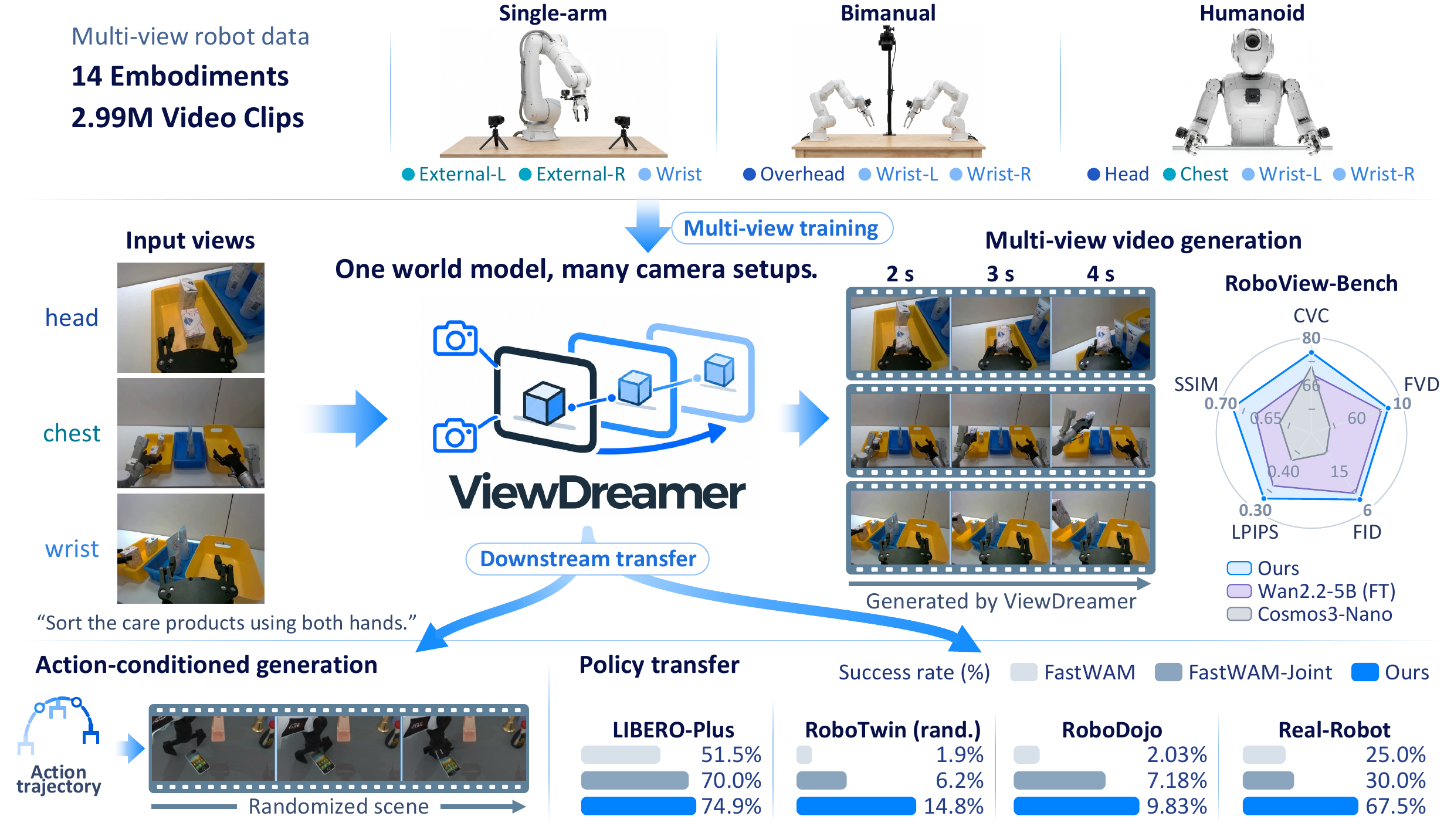}
    \caption{Robots often carry several cameras, whose number and placement vary widely, as in the single-arm, bimanual, and humanoid robots at the top. Pretrained on 2.99M robot video clips from 14 embodiments, \method predicts cross-view-consistent futures for such camera sets from the first frame of every view and a task instruction (middle). It achieves the best three-view scores on RoboView-Bench. Its prior transfers to action-conditioned generation and policy learning (bottom) and improves out-of-domain generalization and camera robustness.}
    \label{fig:teaser}
\end{figure}

\section{Introduction}
\label{section:intro}

Video generation models are increasingly used in robot learning.  They can synthesize training data~\citep{jang2025dreamgen,liu2025robotransfer}, evaluate policies without physical deployment~\citep{zhu2024irasim,guo2025ctrlworld}, and serve as policy backbones~\citep{wu2024gr1,ye2026dreamzero,yuan2026fastwam}.  Robots usually record a task with several synchronized cameras from complementary viewpoints.  An embodied world model should predict what every camera will see, and these predictions must agree wherever views overlap.  We call this property cross-view consistency.

\looseness=-1 Most video generation models, however, produce one video from one image~\citep{blattmann2023videoldm,wan2025}.  A common workaround tiles all views into one grid~\citep{nvidia2026cosmos3,ye2026dreamzero,liu2025robotransfer}.  Tiling trades resolution against compute.  A grid at the pretrained resolution shrinks each view, whereas an enlarged grid makes the computational cost of attention grow quadratically with the view count.  Tiling also ties each camera to a fixed grid position and places the views side by side, a layout rare in single-camera pretraining videos.  Other approaches attend across views or condition on camera geometry yet assume a fixed camera rig~\citep{liao2025genieenvisioner,guo2025ctrlworld,huang2025enerverse,jiang2025evac,paiworld2026}.  Real camera setups differ in number, placement, and order from robot to robot, and even between datasets of one robot.  Bimanual robots typically carry more wrist cameras than single-arm robots, and humanoids carry more cameras than fixed arms~\citep{khazatsky2024droid,agibotworld2025,jiang2025galaxea,wu2025robocoin}.  Embodied world models therefore need a video backbone that stays cross-view consistent across such camera setups.

\looseness=-1 Our key idea is to disentangle per-view generation from cross-view interaction. A pretrained video generator models single-view appearance and motion, yet cannot keep multiple views consistent. We therefore process every view with the same pretrained backbone. Views communicate only through shared \emph{slots}, which are learnable vectors that gather related regions from all views. We train the whole model jointly. The model retains per-view quality and accepts a variable number of views in any order at a computational cost linear in the view count. Without supervision, the slots tend to merge similar patches across views, such as those of one object, yet the grouping stays fragmented. We show that supervising this grouping improves cross-view consistency.

\method implements this design with three components (Fig.~\ref{fig:method-overview}).  After every block of a pretrained video diffusion transformer, a Cross-View Interaction (CVI) module exchanges information among all views through the slots.  Each CVI module starts as an identity mapping, and a View-Count Curriculum (VCC) begins with single views and adds more.  The model thus grows from the pretrained single-view generator into a multi-view one.  Cross-View Slot Agreement (CSA) provides this supervision by distilling region relations from frozen visual foundation models into the slot assignments.  We train \method on 2.99M clips, including real-robot recordings from 14 embodiments, and build \benchname from 500 held-out four-camera episodes.  On \benchname with three and four views, \method leads all compared generators in fidelity and improves cross-view consistency over independent per-view generation.  In a controlled ablation, every component improves policy success and nearly all generation metrics.  Better prediction thus appears to help control.  CVI also produces better videos than full cross-view attention at a lower computational cost.  As an action-conditioned world model, \method scores highest among the compared models in both clean and randomized scenes.  Our policies add CVI to FastWAM-Joint~\citep{yuan2026fastwam} and are initialized from \method.  They outperform the original FastWAM-Joint in simulation with two or three cameras and on a real Franka robot.  In simulation, these gains are largest in out-of-domain (OOD) settings, such as unseen scenes, new camera viewpoints, or sensor noise.

Our contributions are:
\begin{itemize}[leftmargin=1.25em,itemsep=1pt,topsep=2pt]
    \item \method: a robot world model that predicts cross-view-consistent future videos for a variable number of cameras.  It disentangles per-view generation on the pretrained single-view path from cross-view interaction through shared slot tokens.
    \item Cross-View Slot Agreement: a training-time alignment objective that makes geometrically and semantically similar regions in different views share a slot.  A View-Count Curriculum complements it by growing the model from single-view to multi-view generation.
    \item RoboView-Bench: a multi-view world model benchmark where \method achieves the best fidelity and improves cross-view consistency.  Initializing a world-action model and an action-conditioned world model from \method improves OOD generalization.
\end{itemize}

\section{Related Work}
\label{section:related-work}

\parahead{Video generation in robot learning.}
Large video generators capture appearance and motion priors~\citep{blattmann2023videoldm,wan2025}, and interactive models learn controllable dynamics~\citep{yang2023unisim,bruce2024genie}.  Robot learning uses video generators as data engines that synthesize demonstrations~\citep{jang2025dreamgen,liu2025robotransfer,huang2025enerverse,jiang2025evac,guo2025ctrlworld} and as action-conditioned world models (AC-WMs) that evaluate policies without real robots~\citep{zhu2024irasim,rigter2024avid,jiang2025evac,liao2025genieenvisioner,guo2025ctrlworld}.  Video policies extract actions from planned future frames~\citep{du2023unipi}, condition actions on video-model features~\citep{hu2025vpp}, or jointly predict video and actions as world-action models (WAMs)~\citep{wu2024gr1,li2025uva,zhu2025uwm,ye2026dreamzero,yuan2026fastwam,wang2026openwam,zhen2026actionimages}.  We develop a video generator for variable camera sets and validate it in an AC-WM and a WAM.

\parahead{Multi-view robot world models.}
Genie-Envisioner, its simulator GE-Sim 2.0, and Ctrl-World concatenate the tokens of a fixed camera set~\citep{liao2025genieenvisioner,qiu2026gesim2,guo2025ctrlworld}.  Cosmos3, DreamZero, and RoboTransfer tile all views into one frame with a fixed layout~\citep{nvidia2026cosmos3,ye2026dreamzero,liu2025robotransfer}.  EnerVerse, EVAC, PAIWorld, and MVG-WAM condition on calibrated or estimated camera geometry~\citep{huang2025enerverse,jiang2025evac,paiworld2026,chen2026mvgwam}.  These designs require a fixed camera set or camera geometry.  Multi-agent world models support varying view counts.  MultiWorld generates each view from a global 3D-aware state in games and multi-robot simulation~\citep{wu2026multiworld}.  Multics and $\gamma$-World link game players through group or hub tokens~\citep{huang2026multics,liu2026gammaworld}.  These models simulate several agents with separate actions, whereas \method targets one robot with a varying camera set.  Unlike group or hub tokens inside self-attention, our slots sit between backbone blocks, which keeps the pretrained token layout and scales linearly with the view count.  CSA routes each region's patches to one slot for cross-view consistency.

\parahead{Multi-view video generation beyond robotics.}
Conditioned on target camera poses or trajectories, diffusion models generate static objects and scenes~\citep{shi2024mvdream,gao2024cat3d}, novel views across space and time~\citep{watson2025fourdim}, camera-controlled videos~\citep{he2025cameractrl}, and synchronized multi-camera videos~\citep{kuang2024collaborative,xu2024cavia,bai2024syncammaster}.  \method, by contrast, conditions only on each camera's first frame and a task instruction, without camera parameters.  Driving models generate surround-view images or videos from layouts or ego actions~\citep{magicdrive2024,wang2024drivewm,wen2024panacea,gaia2_2025} and typically assume a fixed, outward-facing camera rig.  Robot cameras instead face a shared workspace, and wrist cameras move with the arm.  \method assumes no camera layout and learns which regions correspond across views.

\section{ViewDreamer}
\label{section:method}

\begin{figure*}[t]
    \centering
    \includegraphics[width=\linewidth]{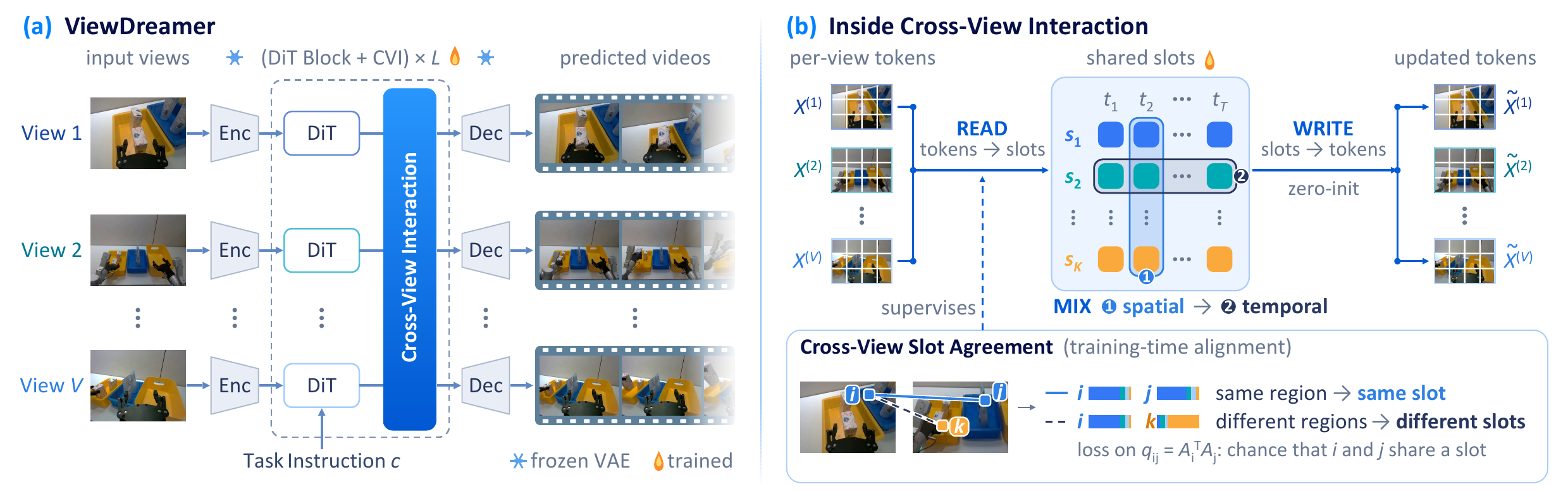}
    \caption{Overview of \method.  (a)~Shared DiT blocks process each view, and a CVI module after every block is the only cross-view path.  (b)~CVI reads all view tokens into shared slots, mixes the slots over space and time, and writes the result back.  CSA aligns the last READ attention with cross-view teacher relations: patches of the same region should share a slot, and patches of different regions should not.}
    \label{fig:method-overview}
\end{figure*}

\subsection{Overview}
\label{section:overview}

Given a task instruction $c$ and the first frame from each of $V$ synchronized cameras, \method predicts future frames for all views. We build on a pretrained single-view video generator, which models individual views well but lacks cross-view consistency. \method therefore disentangles per-view generation from cross-view interaction (Fig.~\ref{fig:method-overview}a). Each view passes independently through $L$ video DiT blocks~\citep{peebles2023dit,wan2025} with shared weights. After each block, a Cross-View Interaction (CVI) module exchanges information through $K$ shared slots: learned tokens that collect features from related regions across views (Sec.~\ref{section:shared-slot}). Our model requires no explicit 3D inputs, such as depth maps or camera parameters, because many robot datasets lack calibration, and off-the-shelf estimates are empirically too inaccurate to improve performance.

We jointly train the DiT blocks and CVI modules on examples with $1$ to $N$ views, with camera number, placement, and order varying across training batches. Cross-View Slot Agreement (CSA, Sec.~\ref{section:slot-agreement}) supervises which patches share a slot, while the View-Count Curriculum (VCC, Sec.~\ref{section:view-curriculum}) gradually increases the number of training views. Together with zero-initialized CVI modules, this curriculum lets the model start from single-view generation and gradually learn cross-view consistency. The base training objective is flow matching~\citep{lipman2023flowmatching}. For view $v$ with clean video latent $z_0^{(v)}$ and Gaussian noise $\epsilon^{(v)}$, the noisy latent at noise level $r$ is $z_r^{(v)}=(1-r)z_0^{(v)}+r\epsilon^{(v)}$. The model learns to predict the velocity $u_r^{(v)}=\epsilon^{(v)}-z_0^{(v)}$. All views within an example share $r$ but use independent noise, allowing them to interact at the same noise level. For each view, the clean first-frame latent replaces the first noisy~frame~as~\mbox{conditioning}. 

\subsection{Cross-View Interaction}
\label{section:shared-slot}

CVI enables cross-view interaction while keeping the token layout expected by the pretrained blocks. Joint self-attention across views or a tiled canvas would alter this layout. CVI instead keeps each view's tokens intact and routes cross-view exchange through $K$ shared slots, a set bottleneck~\citep{lee2019settransformer,jaegle2021perceiver}. At block $\ell$, the DiT processes each view as $X_\ell^{(v)}=\operatorname{Block}_\ell(\tilde{X}_{\ell-1}^{(v)},c,r)$, with its own 3D RoPE grid and first-frame condition. CVI then applies READ, MIX, and WRITE (Fig.~\ref{fig:method-overview}b) to gather features into slots, mix them across slots and time, and return cross-view context to each token.

Let $B$, $T$, and $P$ denote the batch size, latent frame count, and patches per frame. CVI projects the tokens of all views to width $d$ and groups them by latent frame as $H_\ell\in\mathbb{R}^{B\times T\times VP\times d}$. The slots form $S_\ell\in\mathbb{R}^{B\times T\times K\times d}$ and start from learned seeds $S^0\in\mathbb{R}^{K\times d}$ shared across frames. We write $\operatorname{CA}(Q,C)$ for cross-attention from queries $Q$ to context $C$ and $\operatorname{SA}(Z)$ for self-attention within $Z$. Each CVI module computes
\begin{alignat*}{2}
&\text{READ}\qquad & S_\ell^{\mathrm{r}} &= S^0+\operatorname{CA}^{\mathrm{r}}_\ell\bigl(S^0,H_\ell\bigr),\\
&\text{MIX} & S_\ell^{\mathrm{s}} &= S_\ell^{\mathrm{r}}+\operatorname{SA}^{\mathrm{space}}_\ell\bigl(S_\ell^{\mathrm{r}}\bigr),\\
& & S_\ell^{\mathrm{t}} &= S_\ell^{\mathrm{s}}+\operatorname{SA}^{\mathrm{time}}_\ell\bigl(S_\ell^{\mathrm{s}};\Phi_T\bigr),\\
& & S_\ell &= S_\ell^{\mathrm{t}}+\operatorname{MLP}_\ell\bigl(S_\ell^{\mathrm{t}}\bigr),\\
&\text{WRITE} & \tilde{X}_\ell^{(v)} &= X_\ell^{(v)}+\alpha_\ell\,U_\ell\bigl(\operatorname{CA}^{\mathrm{w}}_\ell(H_\ell^{(v)},S_\ell)\bigr).
\end{alignat*}
READ aggregates the tokens of all views into the $K$ slots, independently at each time step. MIX follows factorized space-time attention~\citep{bertasius2021timesformer}. It first applies self-attention across the $K$ slots within each frame and then across the $T$ frames of each slot. The sinusoidal temporal positions $\Phi_T$ enter only the queries and keys. All attention layers and the MLP apply LayerNorm to their inputs.

WRITE lets the tokens of each view, $H_\ell^{(v)}$, attend to the updated slots of the same frame. The zero-initialized output projection $U_\ell$~\citep{zhang2023controlnet} maps this context to the backbone width $D$, and the scalar $\alpha_\ell$ starts at one. The model therefore behaves as the pretrained single-view generator at initialization.

CVI is permutation-equivariant over views, and its computational cost grows linearly with the view count. MIX operates on $K$ slots per frame, regardless of the number of cameras. READ and WRITE add $O(VTPKd)$ per block, compared with $O(V^2T^2P^2D)$ for joint self-attention across views.

\begin{figure*}[t]
    \centering
    \begin{minipage}[t]{0.47\textwidth}
    \vspace{0pt}
    \centering
    \includegraphics[width=0.930\linewidth]{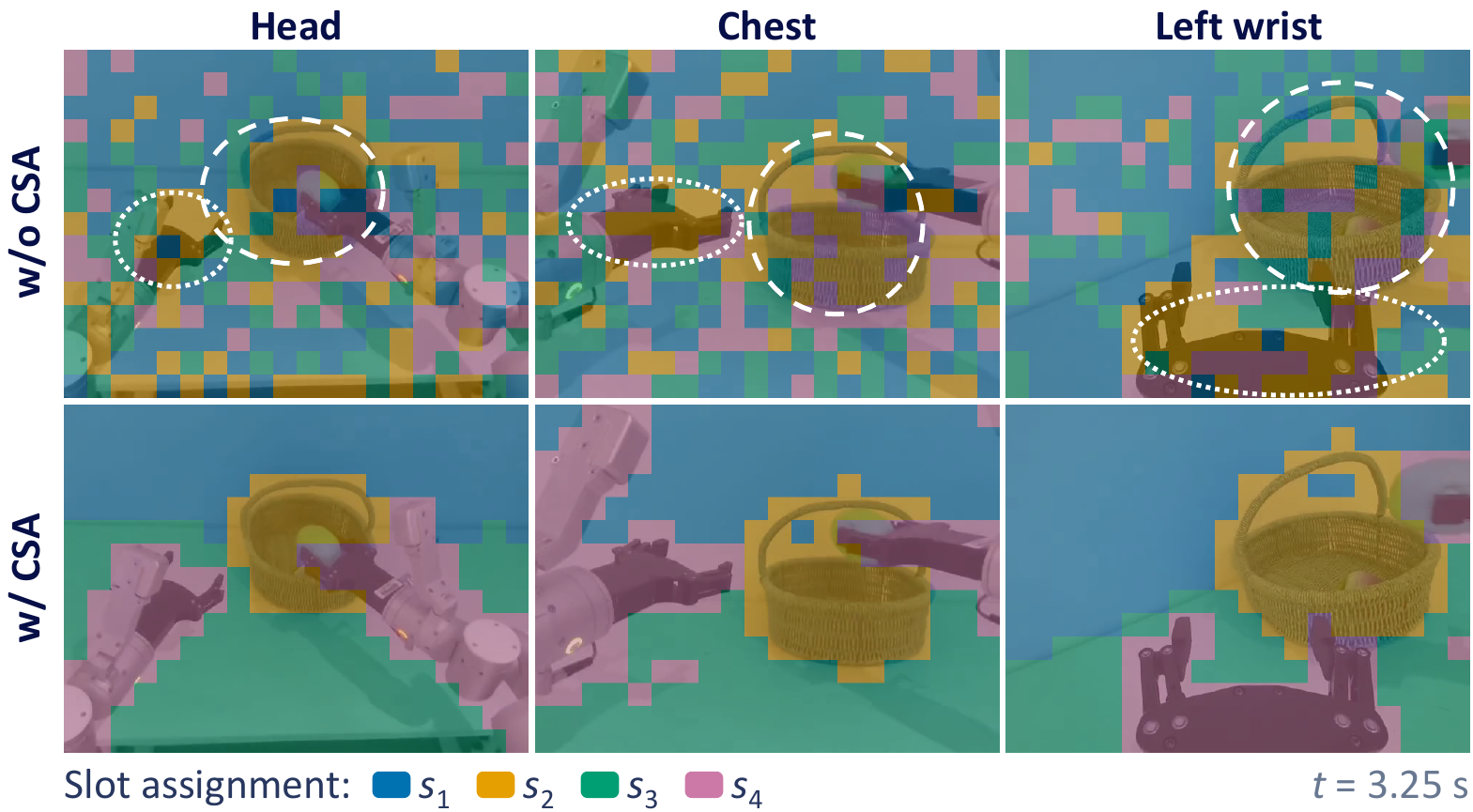}
    \captionof{figure}{Last-block READ slot assignments (colors matched across rows) on a \benchname episode. Without CSA (VCC+CVI), slots partially group regions such as the basket (dashed) and gripper (dotted), but stay fragmented. With CSA (\method), the groups are sharper and more consistent.}
    \label{fig:agreement-attention}
    \end{minipage}\hfill
    \begin{minipage}[t]{0.50\textwidth}
    \vspace{0pt}
    \centering
\setlength{\intextsep}{0pt}\setstretch{1.2}
\begin{algorithm}[H]
\caption{Teacher relations for one latent frame.}
\label{alg:teacher-relations}
\small
\begin{algorithmic}[1]
\REQUIRE unit DINO/VGGT-$\Omega$ descriptors $\mathbf{d}_i,\mathbf{o}_i$, $V\!\geq\!2$, $K$
\ENSURE labels $y$ and cross-view pair sets $\mathcal{P}^{+},\mathcal{P}^{-}$
\STATE $\mathcal{E}\gets\varnothing$; initialize union-find with singleton groups
\FOR{each view pair $a<b$}
\STATE $\sigma_{ij}\gets\frac{1}{2}(\mathbf{d}_i^{\top}\mathbf{d}_j+\mathbf{o}_i^{\top}\mathbf{o}_j)$ for $i\in a$, $j\in b$
\STATE add $(\sigma_{ij},i,j)$ to $\mathcal{E}$ if $i,j$ are mutual nearest neighbors
\ENDFOR
\FOR{$(\sigma_{ij},i,j)\in\mathcal{E}$ in descending $\sigma_{ij}$}
\STATE merge the groups of $i,j$ if distinct and view-disjoint
\ENDFOR
\STATE $\tilde{\mathbf{d}}_i\gets$ normalized mean DINO descriptor of $i$'s group
\STATE $y\gets$ deterministic spherical $K$-means on all $\tilde{\mathbf{d}}_i$
\STATE $\mathcal{P}^{\pm}\gets$ cross-view pairs with $y_i=y_j$ $(+)$ or $y_i\neq y_j$ $(-)$
\end{algorithmic}
\end{algorithm}
    \end{minipage}
\end{figure*}

\subsection{Cross-View Slot Agreement}
\label{section:slot-agreement}

\parahead{Motivation.}\looseness=-1 
Slots decide which patches interact, so patches of one region should share a slot across cameras. Even with long training under flow matching alone, \method tends to group similar patches, but the assignments stay fragmented: a region is often split across slots or merged with its surroundings (Fig.~\ref{fig:agreement-attention}, top). A slot can therefore mix unrelated content into every view. CSA addresses this by supervising slot assignments with teacher relations: a teacher built from frozen visual foundation models links geometrically and semantically similar patches across cameras. Unlike object-centric models~\citep{locatello2020slot,kipf2022savi}, CSA does not tie slots to objects.

\parahead{Teacher relations.}
The teacher matches patches across viewpoints and groups them semantically. For each latent frame, Alg.~\ref{alg:teacher-relations} first matches patches across views with two complementary frozen teachers: DINOv2~\citep{oquab2024dinov2} for semantic similarity and VGGT-$\Omega$~\citep{wang2026vggtomega} for geometric correspondence. Mutual nearest neighbors are merged, by decreasing similarity, into groups of matched patches with at most one patch per view. Spherical $K$-means on the group-averaged DINOv2 features then assigns each patch one of $K$ labels corresponding to the slots. Patch labels are precomputed offline, so neither teacher runs during training or inference.

\parahead{Agreement loss.}
CSA supervises the last-block READ attention to follow the teacher's patch grouping, adding no parameters or inference-time computation. Let $M^{(h)}_{ki}$ be the attention from slot $k$ to patch $i$ in head $h$, normalized over patches. Renormalizing over slots and averaging over $N_{\mathrm{h}}$ heads yields:
\begin{equation*}
    A_{ik}=\frac{1}{N_{\mathrm{h}}}\sum_{h=1}^{N_{\mathrm{h}}}\frac{M^{(h)}_{ki}}{\sum_{k'=1}^{K}M^{(h)}_{k'i}},\qquad \sum_{k=1}^{K}A_{ik}=1.
\end{equation*}
For same-frame patches $i,j$ in different views, $q_{ij}=A_i^\top A_j$ is the probability of sharing a slot. CSA boosts $q_{ij}$ on same-label pairs $\mathcal{P}^{+}$ and lowers it on different-label pairs $\mathcal{P}^{-}$:
\begin{equation*}
    \mathcal{L}_{\mathrm{CSA}}=\frac{1}{|\mathcal{P}^{+}|}\sum_{(i,j)\in\mathcal{P}^{+}}(1-q_{ij})+\frac{1}{|\mathcal{P}^{-}|}\sum_{(i,j)\in\mathcal{P}^{-}}q_{ij},
\end{equation*}
where empty pair sets are skipped. This is a class-balanced $\ell_1$ distance between model and teacher co-association matrices, which only mark same-cluster pairs~\citep{fred2005evidence}. It is thus invariant to slot and label permutations and, unlike set-prediction losses~\citep{carion2020detr}, needs no matching. Pairwise clustering and correspondence losses~\citep{hsu2018kcl,hsu2019mcl,hamilton2022stego}, REPA~\citep{yu2025repa}, and PAIWorld~\citep{paiworld2026} supervise token features, whereas CSA supervises slot assignments. The total training loss is $\mathcal{L}=\mathcal{L}_{\mathrm{flow}}+\lambda_{\mathrm{CSA}}\mathcal{L}_{\mathrm{CSA}}$, where $\mathcal{L}_{\mathrm{flow}}$ is the velocity MSE.

\subsection{View-Count Curriculum}
\label{section:view-curriculum}

VCC stabilizes multi-view training by adding views gradually.  Early steps use many single-view clips to adapt the generator to robot videos.  At step $s$ with horizon $\tau$, examples keep $V_s=1+\operatorname{Binomial}(N-1,p_s)$ uniformly sampled views, with $p_s=\min(s/\tau,1)$, so $\mathbb{E}[V_s]$ rises linearly to $N$.

\section{Experiments}
\label{section:experiments}
\begin{table*}[t]
    \centering
    \small
    \setlength{\tabcolsep}{3.5pt}
    \renewcommand{\arraystretch}{0.96}
    \caption{Multi-view video generation on \benchname with three and four views.  Training never sees four views.  Wan2.2-5B (FT) is fine-tuned on the same data as \method and generates each view independently; the other baselines run zero-shot.  Parentheses give the change from Wan2.2-5B (FT). Best in bold.}
    \label{table:mvwm-bench}
    \begin{tabular*}{\textwidth}{@{\extracolsep{\fill}}lcccccccccc@{}}
        \toprule
        \multirow{2}{*}{Method}
        & \multicolumn{5}{c}{3-view}
        & \multicolumn{5}{c}{4-view} \\
        \cmidrule(lr){2-6} \cmidrule(lr){7-11}
        & SSIM $\uparrow$ & LPIPS $\downarrow$ & FID $\downarrow$ & FVD $\downarrow$ & CVC $\uparrow$
        & SSIM $\uparrow$ & LPIPS $\downarrow$ & FID $\downarrow$ & FVD $\downarrow$ & CVC $\uparrow$ \\
        \midrule
        Wan2.2-5B          & 0.510 & 0.537 & 23.92 & 272.02 & 56.166 & 0.509 & 0.540 & 22.31 & 259.90 & 56.284 \\
        Wan2.1-14B         & 0.484 & 0.570 & 24.99 & 261.73 & 58.314 & 0.483 & 0.572 & 23.75 & 254.20 & 58.715 \\
        Wan2.2-A14B        & 0.495 & 0.559 & 20.84 & 247.55 & 55.341 & 0.491 & 0.566 & 20.06 & 253.16 & 53.640 \\
        Genie-Envisioner   & 0.686 & 0.420 & 33.20 & 184.73 & 71.750 & 0.657 & 0.452 & 61.28 & 429.01 & 71.100 \\
        Cosmos3-Edge       & 0.590 & 0.530 & 40.85 & 392.89 & 53.760 & 0.646 & 0.460 & 28.42 & 309.27 & 65.128 \\
        Cosmos3-Nano       & 0.635 & 0.429 & 19.25 & 89.57 & 71.632 & 0.675 & 0.379 & 13.71 & 88.16 & \textbf{77.714} \\
        \midrule
        Wan2.2-5B (FT)     & 0.662 & 0.363 & 9.92 & 32.93 & 69.316 & 0.658 & 0.371 & 8.58 & 29.50 & 69.901 \\
        \textsf{\bfseries Ours}     & \improve{\textbf{0.687}}{+0.025} & \improve{\textbf{0.330}}{$-0.033$} & \improve{\textbf{8.47}}{$-1.45$} & \improve{\textbf{25.41}}{$-7.52$} & \improve{\textbf{75.693}}{+6.377} & \improve{\textbf{0.688}}{+0.030} & \improve{\textbf{0.332}}{$-0.039$} & \improve{\textbf{7.13}}{$-1.45$} & \improve{\textbf{21.26}}{$-8.24$} & \improve{75.150}{+5.249} \\
        \bottomrule
    \end{tabular*}
\end{table*}

\subsection{Experimental Setting}
\label{section:experimental-setting}

\begin{figure*}[t]
    \centering
    \includegraphics[width=\linewidth]{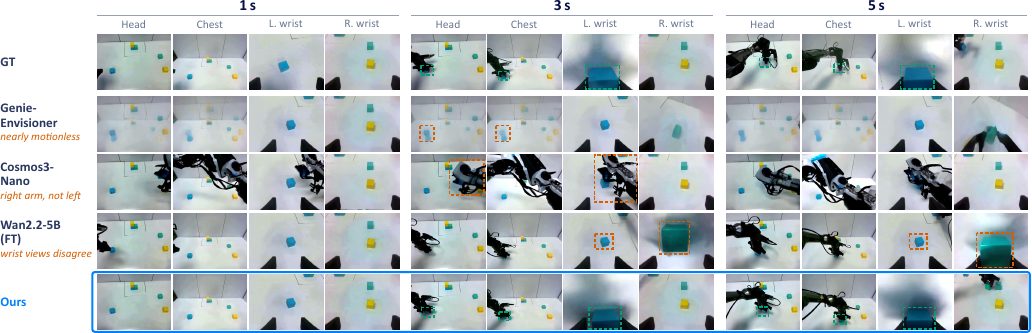}
    \caption{Four-view generation on \benchname (task: ``Put the blue build block in the center of the table with left arm.''); training never sees four views.  Rows are methods; column groups show all views at 1, 3, and 5\,s.  Orange notes and boxes mark errors.  Genie-Envisioner is nearly motionless, Cosmos3-Nano moves the right arm instead of the left, and the wrist views of Wan2.2-5B (FT) disagree with its other views.  Only \method keeps all views consistent, as in the ground truth (green boxes).  More examples are in App.~\ref{section:appendix-generation}.}
    \label{fig:qualitative-results}
\end{figure*}
\parahead{Tasks.}
We evaluate \method on multi-view video generation, action-conditioned video generation, and world-action policy learning.  For the latter two, \method initializes an action-conditioned world model (AC-WM) and a world-action model (WAM).

\begin{table*}[t]
    \centering
    \small
    \setlength{\tabcolsep}{4pt}
    \caption{Action-conditioned generation on RoboTwin 2.0 with three views (higher is better).  Clean (IID): 250 held-out episodes; Random (OOD): 250 randomized episodes of the same tasks.  EWMScore~\citep{shang2026worldarena} is the mean of 15 metrics.  Offsets compare Ours with Wan2.2-5B.}
    \label{table:worldarena}
    \begin{tabular*}{\textwidth}{@{\extracolsep{\fill}}lccccccc@{}}
        \toprule
        Method & EWMScore & \shortstack{Visual\\quality} & \shortstack{Motion\\quality} & \shortstack{Content\\consistency} & \shortstack{Physics\\adherence} & \shortstack{3D\\accuracy} & Controllability \\
        \midrule
        \multicolumn{8}{@{}l}{\textit{Clean (IID)}} \\
        Cosmos3-Edge & 47.70 & 47.63 & 35.98 & \textbf{52.51} & 20.28 & 75.86 & 57.43 \\
        Cosmos3-Nano & \underline{49.45} & 50.00 & 37.62 & \underline{51.99} & \textbf{23.67} & \textbf{78.61} & \textbf{59.17} \\
        \cmidrule(l{0pt}r{0pt}){1-8}
        Wan2.2-5B & 48.33 & \underline{51.72} & \underline{38.41} & 51.41 & 18.01 & 75.92 & 56.20 \\
        \textsf{\bfseries Ours} & \improve{\textbf{50.09}}{+1.76} & \improve{\textbf{54.72}}{+3.00} & \improve{\textbf{40.74}}{+2.33} & \degrade{50.58}{$-0.83$} & \improve{\underline{21.19}}{+3.18} & \improve{\underline{77.34}}{+1.42} & \improve{\underline{58.11}}{+1.91} \\
        \midrule
        \multicolumn{8}{@{}l}{\textit{Random (OOD)}} \\
        Cosmos3-Edge & 25.08 & 17.14 & 21.66 & 13.81 & 10.30 & 47.89 & 51.01 \\
        Cosmos3-Nano & 31.39 & 29.36 & 26.78 & 19.77 & 11.38 & 58.36 & 51.79 \\
        \cmidrule(l{0pt}r{0pt}){1-8}
        Wan2.2-5B & \underline{43.08} & \underline{36.24} & \underline{40.41} & \textbf{53.69} & \underline{13.08} & \textbf{61.56} & \underline{52.98} \\
        \textsf{\bfseries Ours} & \improve{\textbf{46.99}}{+3.91} & \improve{\textbf{42.51}}{+6.27} & \improve{\textbf{54.81}}{+14.40} & \degrade{\underline{52.73}}{$-0.96$} & \improve{\textbf{13.30}}{+0.22} & \degrade{\underline{61.05}}{$-0.51$} & \improve{\textbf{53.02}}{+0.04} \\
        \bottomrule
    \end{tabular*}
\end{table*}

\parahead{Data.}
We train \method on 2,990,654 five-second clips from AgiBotWorld-Beta~\citep{agibotworld2025}, Galaxea~\citep{jiang2025galaxea}, RoboCOIN~\citep{wu2025robocoin}, RoboMIND~\citep{wu2025robomind}, and RoVid-X~\citep{deng2026revidgen}.  Real-robot clips cover 14 embodiments, 12 of them with multiple views.  Curation removes low-motion clips and invalid captions.  The training set retains up to three views per clip and excludes all \benchname episodes.

\parahead{Baselines.}\looseness=-1 
The multi-view generation baselines are the Wan models~\citep{wan2025}, Genie-Envisioner~\citep{liao2025genieenvisioner}, and Cosmos3~\citep{nvidia2026cosmos3}, all run zero-shot from official checkpoints.  Wan2.2-5B (FT) is fine-tuned on our data at our batch size and generates each view independently.  For AC-WMs, we adapt Wan2.2-5B with our recipe and fine-tune both Cosmos3 models with the same batch size and number of epochs.  The policy baselines are the VLAs $\pi_0$, $\pi_{0.5}$~\citep{black2024pi0,black2025pi05}, StarVLA~\citep{starvla2026}, and ABot-M0~\citep{abot2026}, and the WAMs OpenWAM-$\alpha$~\citep{wang2026openwam}, FastWAM, and FastWAM-Joint~\citep{yuan2026fastwam}.  FastWAM-Joint's action queries attend to all video tokens, whereas FastWAM's see only the first frame.

\parahead{Implementation details.}
We initialize \method from Wan2.2-TI2V-5B (Wan2.2-5B)~\citep{wan2025} and train the DiT and CVI modules with the VAE and text encoder frozen. CVI uses $K=4$ slots of width $d=256$ and four attention heads.  It adds 86.8M parameters, under 2\% of the backbone.  We set $\lambda_{\mathrm{CSA}}=0.1$.  Each view has 81 frames of $640\times480$ at 16\,FPS, which yields a $21\times15\times20$ grid of DiT tokens.  We use AdamW with a learning rate of $10^{-5}$, a weight decay of $0.01$, 500 warmup steps, and cosine decay.  We train \method and Wan2.2-5B (FT) for 32K steps on 128 B200 GPUs with a global batch size of 1024.  Ablation runs use 20K steps at a batch size of 256.  \method and the Wan baselines sample \benchname videos with 40 Euler steps, a guidance scale of 5, and a shift of 5.

\subsection{Multi-View Video Generation}
\label{section:mv-generation}

\parahead{\benchname.}
We introduce \benchname to evaluate multi-view generation on 500 held-out RoboCOIN episodes covering 211 tasks across multiple robots. Camera configurations vary by robot and include wrist-mounted, body-mounted, and external cameras. We evaluate each episode in both three- and four-view settings. Training uses at most \(N=3\) views per clip, and footage from the fourth RoboCOIN cameras is reserved for four-view evaluation. This setting tests generalization to both an unseen view count and held-out camera placements.

\parahead{Setting.}
Given the first frame of every view and the instruction, each model predicts the remaining 80 frames, evaluated at $640\times480$.  SSIM and LPIPS measure paired fidelity, and FID and FVD~\citep{unterthiner2018fvd} compare distributions.  Our cross-view consistency (CVC) score adapts MEt3R~\citep{asim2025met3r} to simultaneous frames of two cameras.  We use every camera pair at frames $t\in\{10,20,\dots,80\}$, resized to $256\times256$.  MEt3R returns a feature dissimilarity and an overlap fraction: $(d_{\mathrm{g}},o_{\mathrm{g}})$ for the generated pair and $(d_{\mathrm{r}},o_{\mathrm{r}})$ for the real pair.  Each camera pair and frame with real overlap $o_{\mathrm{r}}\geq0.1$ scores
\begin{equation*}
    100\sqrt{\min\bigl(1,\tfrac{d_{\mathrm{r}}+\delta}{d_{\mathrm{g}}+\delta}\bigr)\min\bigl(1,\tfrac{o_{\mathrm{g}}}{o_{\mathrm{r}}}\bigr)},\qquad\delta=10^{-6},
\end{equation*}
which equals 100 when the generated pair is no more dissimilar and overlaps no less than the real pair.  CVC takes the median over each pair's frames and the mean over pairs and episodes, and it needs no camera poses.

\parahead{Results.}
\method achieves the best SSIM, LPIPS, FID, and FVD of all compared generators with three and four views, and the best three-view CVC (Tab.~\ref{table:mvwm-bench}).  It also beats Wan2.2-5B (FT), trained on the same data, on every metric, and its fidelity gain grows from one to four views (Tab.~\ref{table:fewer-views}).  Even with a single view, \method improves slightly.  A VLM judge also rates its three-view consistency highest (Tab.~\ref{table:vlm-judge}).  Wan2.2-5B (FT) generates each view alone, so its wrist and head views can disagree on where an object is (Fig.~\ref{fig:qualitative-results}).  With four views, \method keeps a clear lead in fidelity, whereas Genie-Envisioner, trained only with three camera roles, more than doubles its FVD.  Cosmos3-Nano reaches the highest four-view CVC, yet its FVD is more than four times ours.  Nearly every method predicts wrist views worst, and \method gains most there, cutting wrist LPIPS by 0.050 over Wan2.2-5B (FT) versus 0.019 elsewhere (Tab.~\ref{table:camera-type}).  Other cameras likely supply context missing from fast-moving wrist views.

\subsection{Action-Conditioned Video Generation}
\label{section:ac-generation}

\parahead{Setting.}
We fine-tune \method and the official Wan2.2-5B into AC-WMs with the same EVAC-style recipe~\citep{jiang2025evac}.  \method keeps its CVI modules, and Wan2.2-5B attends jointly over all views.  Training uses 2,250 clean RoboTwin 2.0 episodes, 45 per task.  Clean (IID) evaluation uses the 250 held-out clean episodes, and Random (OOD) evaluation uses 250 randomized-scene episodes of the same tasks.  EWMScore~\citep{shang2026worldarena} rates all three views.

\parahead{Results.}
Our AC-WM obtains the highest EWMScore on both splits (Tab.~\ref{table:worldarena}).  Its margin over Wan2.2-5B more than doubles under randomization, from 1.76 to 3.91 points.  It also degrades least from Clean to Random, losing 3.10 EWMScore points versus 5.25 for Wan2.2-5B.  The Cosmos3 models, which generate wrist views at a lower resolution, degrade far more.  This robustness suggests that the \method prior is less tied to the appearance of the fine-tuning scenes.  The prior improves visual and motion quality on both splits.  On Clean, it also improves physics adherence, 3D accuracy, and controllability.

\begin{table*}[t]
    \centering
    \small
    \setlength{\tabcolsep}{1.8pt}
    \caption{Single-arm policy success (\%).  LIBERO tests in-distribution scenes; LIBERO-Plus tests the same policies zero-shot under seven perturbations.  Ours builds on FastWAM-Joint; offsets are relative to it (green: better; red: worse).  Best in bold, second best underlined.}
    \label{table:policy-model-comparison}
    \begin{tabular*}{\textwidth}{@{\extracolsep{\fill}}lccccccccc@{}}
        \toprule
        \multirow{2}{*}{Method}
        & \multirow{2}{*}{\shortstack{LIBERO\\Total $\uparrow$}}
        & \multicolumn{8}{c}{LIBERO-Plus} \\
        \cmidrule(lr){3-10}
        & & Camera $\uparrow$ & Robot $\uparrow$ & Language $\uparrow$ & Light $\uparrow$
        & Background $\uparrow$ & Noise $\uparrow$ & Layout $\uparrow$ & Total $\uparrow$ \\
        \midrule
        $\pi_0$ & 94.4 & 13.8 & 6.0 & 58.8 & 85.0 & 81.4 & 79.0 & 68.9 & 53.6 \\
        $\pi_{0.5}$ & 96.9 & \textbf{78.4} & \underline{73.6} & 80.8 & 96.2 & \underline{94.1} & \textbf{89.0} & \textbf{84.5} & \textbf{84.4} \\
        StarVLA & 96.6 & 52.5 & 49.8 & 88.5 & 95.7 & \textbf{95.7} & 73.0 & 76.9 & 74.1 \\
        ABot-M0 & \underline{98.6} & \underline{60.4} & 67.9 & 86.4 & 96.2 & 91.6 & \underline{86.4} & \underline{82.6} & \underline{80.5} \\
        FastWAM & 97.6 & 16.4 & 44.5 & 68.9 & 78.2 & 53.7 & 37.7 & 60.7 & 51.5 \\
        OpenWAM-$\alpha$ & \textbf{99.3} & 33.8 & \textbf{76.1} & 88.0 & \textbf{97.0} & 87.1 & 39.8 & 77.5 & 69.2 \\
        \midrule
        FastWAM-Joint & 98.5 & 40.7 & 71.5 & \textbf{94.7} & \underline{96.3} & 58.0 & 55.3 & 78.8 & 70.0 \\
        \textsf{\bfseries Ours} &\improve{\underline{98.6}}{+0.1} & \improve{53.6}{+12.9} & \same{71.5}{0.0} & \degrade{\underline{93.3}}{$-1.4$} & \degrade{94.7}{$-1.6$} & \improve{63.6}{+5.6} & \improve{69.1}{+13.8} & \improve{81.0}{+2.2} & \improve{74.9}{+4.9} \\
        \bottomrule
    \end{tabular*}
\end{table*}

\begin{table}[t]
    \centering
    \begin{minipage}{0.62\textwidth}
    \centering
    \small
    \setlength{\tabcolsep}{1.2pt}
    \caption{Bimanual success (\%) on RoboTwin 2.0 and RoboDojo.  Clean2Random trains on clean demonstrations only and tests on clean and randomized scenes.  Ours builds on FastWAM-Joint, and offsets are relative to it.  Best in bold, second best underlined.}
    \label{table:bimanual-policy-comparison}
    \begin{tabular*}{\linewidth}{@{\extracolsep{\fill}}lccccc@{}}
        \toprule
        \multirow{2}{*}{Method}
        & \multicolumn{3}{c}{RoboTwin 2.0}
        & \multicolumn{2}{c}{RoboDojo} \\
        \cmidrule(lr){2-4} \cmidrule(l){5-6}
        & Clean & Rand. & Avg & SR & Score \\
        \midrule
        $\pi_0$ & 46.4 & 16.3 & 31.4 & 1.53 & 3.48 \\
        $\pi_{0.5}$ & 70.7 & \underline{46.0} & \underline{58.4} & 6.93 & 11.44 \\
        StarVLA & 46.5 & 3.2 & 24.9 & 3.24 & 6.40 \\
        ABot-M0 & 70.7 & 36.0 & 53.4 & 1.73 & 3.67 \\
        FastWAM & 77.8 & 1.9 & 39.9 & 2.03 & 3.48 \\
        OpenWAM-$\alpha$ & \textbf{89.4} & \textbf{48.7} & \textbf{69.0} & \textbf{11.92} & \textbf{17.18} \\
        \midrule
        FastWAM-Joint & 80.9 & 6.2 & 43.5 & 7.18 & 11.35 \\
        \textsf{\bfseries Ours} & \improve{\underline{83.8}}{+2.9} & \improve{14.8}{+8.6} & \improve{49.3}{+5.8} & \improve{\underline{9.83}}{+2.65} & \improve{\underline{15.24}}{+3.89} \\
        \bottomrule
    \end{tabular*}
    \end{minipage}
\end{table}

\begin{table}[t]
    \centering
    \begin{minipage}{0.62\textwidth}
    \centering
    \small
    \setlength{\tabcolsep}{1.2pt}
    \renewcommand{\arraystretch}{1.25}
    \caption{Real-robot success rate / score (\%) over 10 trials per task.  Best in bold, second best underlined.}
    \label{table:real-robot-results}
    \begin{tabular*}{\linewidth}{@{\extracolsep{\fill}}lccccc@{}}
        \toprule
        Method & \shortstack{Cup\vphantom{p}\\into box\vphantom{b}} & \shortstack{Cubes\vphantom{p}\\into cups\vphantom{b}} & \shortstack{Stack\vphantom{p}\\cups\vphantom{b}} & \shortstack{Stack\vphantom{p}\\cubes\vphantom{b}} & Avg. \\
        \midrule
        FastWAM & 40\,/\,-- & \underline{30}\,/\,35 & \underline{10}\,/\,\underline{35} & 20\,/\,\underline{40} & 25.0\,/\,\underline{36.7} \\
        FastWAM-Joint & \underline{60}\,/\,-- & 20\,/\,\underline{40} & \underline{10}\,/\,30 & \underline{30}\,/\,\underline{40} & \underline{30.0}\,/\,\underline{36.7} \\
        \midrule
        \textsf{\bfseries Ours} & \textbf{80}\,/\,-- & \textbf{90}\,/\,\textbf{95} & \textbf{40}\,/\,\textbf{65} & \textbf{60}\,/\,\textbf{75} & \textbf{67.5}\,/\,\textbf{78.3} \\
        \bottomrule
    \end{tabular*}
    \end{minipage}
\end{table}

\subsection{World-Action Policy Transfer}
\label{section:policy-transfer}

\parahead{Setting.}
Each WAM predicts an action chunk and future visual latents from the current multi-view observations.  Our WAM adds CVI to FastWAM-Joint and initializes its video network (and, in simulation, its action network) from \method.  On LIBERO-Plus~\citep{fei2025liberoplus}, we test the two-camera LIBERO policies zero-shot on all 10,030 instances of its seven perturbation types.  RoboTwin 2.0~\citep{chen2025robotwin2} has 50 three-camera bimanual tasks.  Its Clean2Random setting trains on clean demonstrations and tests on 100 clean and 100 randomized episodes per task.  The bimanual RoboDojo benchmark~\citep{chen2026robodojo} reports success rate (SR) and score.

Real-robot tests use a Franka in the DROID setup~\citep{khazatsky2024droid} with two external cameras and one wrist camera.  The robot puts a cup into a box, places two cubes into cups of matching color, and stacks three cups or three cubes.  Each policy learns all four tasks from 50 demonstrations each, and test-time object positions usually differ somewhat from the demonstrations.  Multi-step tasks also get a score that counts one of two subtasks as half a success.  Its average covers these three tasks.  See App.~\ref{section:appendix-real-robot} for details.

\parahead{Simulation results.}
Compared with FastWAM-Joint, our WAM mainly improves OOD generalization (Tabs.~\ref{table:policy-model-comparison} and~\ref{table:bimanual-policy-comparison}).  Both policies saturate standard LIBERO~\citep{liu2023libero} (98.5\% and 98.6\%); on LIBERO-Plus, our WAM raises success from 70.0\% to 74.9\%.  The largest gains come under sensor noise and camera viewpoint changes.  The camera gain grows with the shift.  When the third-person camera orbits the scene by more than 30$^\circ$, success rises from 24.6\% to 44.1\% (Fig.~\ref{fig:libero-plus-camera}).  Background and layout also improve.  On RoboTwin 2.0, our WAM more than doubles randomized success, from 6.2\% to 14.8\%, and improves clean success.  It ranks second on RoboDojo.  The policies that lead on some benchmarks, $\pi_{0.5}$, ABot-M0, and OpenWAM-$\alpha$, pretrain on thousands of hours of embodied data with actions, whereas our pretraining uses only videos, without actions.

\parahead{Real-robot results.}
Our WAM outperforms both FastWAM variants on every real-robot task and raises the average success rate from 30.0\% (FastWAM-Joint) to 67.5\% (Tab.~\ref{table:real-robot-results}).  The gain is largest on cubes into cups, from 20\% to 90\%.  On the two stacking tasks, our WAM doubles or quadruples the success of FastWAM-Joint (60\% vs.\ 30\% for cubes and 40\% vs.\ 10\% for cups).  It also completes more subtasks: its average score over the three multi-step tasks rises from 36.7\% to 78.3\%.  All three policies use the same 200 demonstrations, the same three cameras, and the same trial protocol.

\begin{table*}[t]
    \centering
    \footnotesize
    \setlength{\tabcolsep}{1.0pt}
    \caption{Component ablation on \benchname (ablation recipe).  LIBERO-Plus SR uses the same checkpoints.  Best in bold.}
    \label{table:main-ablation}
    \begin{tabular*}{\textwidth}{@{\extracolsep{\fill}}cccccccccccccc@{}}
            \toprule
            \multicolumn{3}{c}{Modules}
            & \multicolumn{5}{c}{3-view}
            & \multicolumn{5}{c}{4-view}
            & \multirow{2}{*}{\shortstack{LIBERO-\\Plus SR $\uparrow$}} \\
            \cmidrule(lr){1-3} \cmidrule(lr){4-8} \cmidrule(lr){9-13}
            VCC & CVI & CSA
            & SSIM $\uparrow$ & LPIPS $\downarrow$ & FID $\downarrow$ & FVD $\downarrow$ & CVC $\uparrow$
            & SSIM $\uparrow$ & LPIPS $\downarrow$ & FID $\downarrow$ & FVD $\downarrow$ & CVC $\uparrow$ & \\
            \midrule
            \multicolumn{3}{c}{Independent views} & 0.653 & 0.376 & 10.52 & 36.32 & 67.367 & 0.649 & 0.384 & 9.18 & 31.79 & 67.323 & 69.00 \\
            & $\checkmark$ & & 0.663 & 0.364 & 9.93 & 32.36 & 69.527 & 0.662 & 0.367 & 8.59 & 29.32 & 70.150 & 72.64 \\
            & $\checkmark$ & $\checkmark$ & 0.672 & 0.351 & 9.69 & 31.30 & 70.538 & 0.670 & 0.359 & 8.44 & 28.38 & 71.065 & 73.89 \\
            $\checkmark$ & & & 0.660 & 0.366 & 10.34 & 34.18 & 68.438 & 0.657 & 0.373 & 9.07 & 29.75 & 68.052 & 69.37 \\
            $\checkmark$ & $\checkmark$ & & 0.670 & 0.355 & 9.43 & \textbf{29.55} & 71.148 & 0.667 & 0.362 & 8.10 & \textbf{26.38} & 70.999 & 73.15 \\
            $\checkmark$ & $\checkmark$ & $\checkmark$ & \textbf{0.675} & \textbf{0.348} & \textbf{9.25} & 30.59 & \textbf{72.630} & \textbf{0.672} & \textbf{0.353} & \textbf{7.91} & 26.98 & \textbf{73.178} & \textbf{74.18} \\
            \bottomrule
    \end{tabular*}
\end{table*}

\begin{table*}[t]
    \centering
    \begin{minipage}[b]{0.58\textwidth}
    \centering
    \footnotesize
    \setlength{\tabcolsep}{1.2pt}
    \captionof{table}{Cross-view designs from Wan2.2-TI2V-5B, all with VCC and no CSA (ablation recipe).  Canvas tiles native-size views under one RoPE grid.  Full attn.\ keeps per-view RoPE.  Best in bold.}
    \label{table:design-comparison}
    \begin{tabular*}{\linewidth}{@{\extracolsep{\fill}}lcccccc@{}}
        \toprule
        \multirow{2}{*}{}
          & \multicolumn{3}{c}{3-view}
          & \multicolumn{3}{c}{4-view} \\
        \cmidrule(lr){2-4}\cmidrule(lr){5-7}
          & SSIM $\uparrow$ & FVD $\downarrow$ & CVC $\uparrow$
          & SSIM $\uparrow$ & FVD $\downarrow$ & CVC $\uparrow$ \\
        \midrule
        Canvas         & 0.658 & 35.32 & 65.210 & 0.648 & 30.38 & 67.800 \\
        Full attn.     & 0.645 & 37.48 & 67.380 & 0.639 & 33.58 & 68.138 \\
        CVI            & \textbf{0.670} & \textbf{29.55} & \textbf{71.148} & \textbf{0.667} & \textbf{26.38} & \textbf{70.999} \\
        \bottomrule
    \end{tabular*}
    \par\hrule height 0pt
    \end{minipage}\hfill
    \begin{minipage}[b]{0.40\textwidth}
    \centering
    \footnotesize
    \setlength{\tabcolsep}{1.2pt}
    \captionof{table}{Effect of the slot count $K$ (shaded: full model). The other settings follow the full model (ablation recipe). Best in bold.}
    \label{table:slot-number-ablation}
    \begin{tabularx}{\linewidth}{@{}l*{6}{>{\centering\arraybackslash}X}@{}}
        \toprule
        \multirow{2}{*}{$K$}
          & \multicolumn{3}{c}{3-view}
          & \multicolumn{3}{c}{4-view} \\
        \cmidrule(lr){2-4}\cmidrule(lr){5-7}
          & SSIM $\uparrow$ & FVD $\downarrow$ & CVC $\uparrow$
          & SSIM $\uparrow$ & FVD $\downarrow$ & CVC $\uparrow$ \\
        \midrule
        2 & 0.677 & 32.92 & 70.244 & \textbf{0.677} & 28.97 & 69.814 \\
        \oursrow 4 & 0.675 & \textbf{30.59} & \textbf{72.630}
          & 0.672 & \textbf{26.98} & \textbf{73.178} \\
        8 & \textbf{0.678} & 33.71 & 71.815 & 0.676 & 28.56 & 71.445 \\
        \bottomrule
    \end{tabularx}
    \par\hrule height 0pt
    \end{minipage}
\end{table*}

\begin{figure*}[t]
    \centering
    \begin{minipage}[b]{0.55\textwidth}
    \centering
    \footnotesize
    \setlength{\tabcolsep}{1.2pt}
    \captionof{table}{Effect of $\lambda_{\mathrm{CSA}}$ and the teacher (shaded: full model). All runs use VCC and CVI (ablation recipe). w/o~VGGT-$\Omega$: DINO-only teacher, $\lambda_{\mathrm{CSA}}=0.1$. Best in bold.}
    \label{table:agreement-weight-ablation}
    \begin{tabularx}{\linewidth}{@{}l*{6}{>{\centering\arraybackslash}X}@{}}
        \toprule
        \multirow{2}{*}{$\lambda_{\mathrm{CSA}}$}
          & \multicolumn{3}{c}{3-view}
          & \multicolumn{3}{c}{4-view} \\
        \cmidrule(lr){2-4}\cmidrule(lr){5-7}
          & SSIM $\uparrow$ & FVD $\downarrow$ & CVC $\uparrow$
          & SSIM $\uparrow$ & FVD $\downarrow$ & CVC $\uparrow$ \\
        \midrule
        0 & 0.670 & \textbf{29.55} & 71.148
          & 0.667 & \textbf{26.38} & 70.999 \\
        \oursrow 0.1 & 0.675 & 30.59 & 72.630
          & 0.672 & 26.98 & \textbf{73.178} \\
        0.5 & 0.678 & 35.09 & \textbf{72.995}
          & 0.677 & 30.65 & 72.532 \\
        1.0 & \textbf{0.680} & 40.92 & 69.713
          & \textbf{0.679} & 36.22 & 70.703 \\
        \midrule
        w/o VGGT-$\Omega$ & 0.672 & 31.14 & 72.088 & 0.669 & 26.74 & 71.639 \\
        \bottomrule
    \end{tabularx}
    \par\hrule height 0pt
    \end{minipage}\hfill
    \begin{minipage}[b]{0.42\textwidth}
    \centering
    \includegraphics[width=0.949\linewidth,trim=0 0 0 4.5pt,clip]{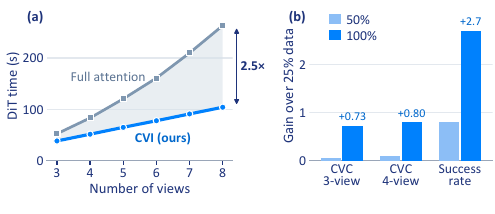}
    \captionof{figure}{(a)~DiT GPU time per sample (80 forward passes, no VAE) on one B200.  (b)~Gains in three- and four-view CVC and LIBERO-Plus success over the nested 25\% subset (ablation recipe).}
    \label{fig:latency-scaling}
    \par\hrule height 0pt
    \end{minipage}
\end{figure*}
\subsection{Ablation Study}
\label{section:ablations}

Tab.~\ref{table:main-ablation} reports runs of 20K steps at a global batch size of 256.  Each checkpoint initializes a WAM fine-tuned on LIBERO as in Sec.~\ref{section:policy-transfer}.  Every component improves SSIM, LPIPS, FID, and CVC, and the full model is best on LIBERO-Plus.

\parahead{Cross-View Interaction.}
Adding CVI to the independent-view baseline improves policy success and every generation metric; three-view FVD drops from 36.32 to 32.36.  Given the same views and data and the small single-view gain (Tab.~\ref{table:fewer-views}), most of the gain likely comes from cross-view exchange.  Without CVI in pretraining or policy, LIBERO-Plus success is 69.0\%, close to FastWAM-Joint's 70.0\%.

\parahead{Cross-view design.}
Tab.~\ref{table:design-comparison} compares CVI with canvas tiling and full self-attention, which also use no view embedding.  Both alternatives trail CVI on every reported metric, likely because they disturb the token layout of the pretrained blocks.  CVI also requires less computation.  Each extra view adds a nearly constant 13\,s of DiT time per sample, whereas full-attention time grows superlinearly (Fig.~\ref{fig:latency-scaling}a).

\parahead{View-Count Curriculum.}
Runs without VCC use all native views of every clip.  VCC helps every variant and lowers the three-view FVD of CVI from 32.36 to 29.55.  It helps even without CVI.  With CVI, it lets cross-view interaction grow out of the single-view solution.

\parahead{Cross-View Slot Agreement.}
CSA improves the SSIM, LPIPS, FID, and CVC of VCC+CVI and raises LIBERO-Plus success from 73.2\% to 74.2\%.  Its CVC gain grows from 1.5 points with three views to 2.2 with four, where more views must agree.  CSA also sharpens the slot assignments (Fig.~\ref{fig:agreement-attention}).  Without CSA, a region's patches spread over several slots.  With CSA, the same-slot share of adjacent patches rises from 39\% to 87\% (27\% and 42\% if shuffled).

\parahead{Slot count, agreement weight, and teacher.}\looseness=-1 
Four slots give the best FVD and CVC with both view counts (Tab.~\ref{table:slot-number-ablation}).  A DINO-only teacher (Tab.~\ref{table:agreement-weight-ablation}, last row) improves CVC over no CSA yet trails our teacher by 0.5 and 1.5 points with three and four views.  Geometric correspondences may matter more as the view count grows.

\parahead{Data scaling.}
CVC and LIBERO-Plus success improve with more training data, mostly from half to full data (Fig.~\ref{fig:latency-scaling}b).  Even with a quarter of the data, the full model beats the all-data independent-view baseline in CVC (71.9 vs.\ 67.4) and LIBERO-Plus success (71.5\% vs.\ 69.0\%).

\section{Conclusion}
\label{section:conclusion}

We introduce \method, a novel multi-view robot world model that improves consistency across views by running a pretrained single-view generator per camera and linking cameras through shared slots.  Extensive evaluations show the best generation quality and stronger downstream policies in simulation and on a real robot, underscoring the potential of one model for robots with diverse camera setups.  The gains are largest out of domain, including under new camera viewpoints, scenes, and sensor noise.

\parahead{Limitations and Future Work.}
Generating consistent wrist-camera views remains challenging, as these cameras move quickly. Our design restricts neither the backbone nor the camera setup, yet truly arbitrary setups remain unverified.  Our experiments cover a limited range of viewpoints and camera types, at most four cameras, and only \mbox{Wan2.2-5B}.  Broader placements, more cameras, and larger backbones remain to be tested, and future work could extend \method to multi-agent games and autonomous driving.

\clearpage
\newpage
\bibliographystyle{assets/plainnat}
\bibliography{paper}

\clearpage
\newpage
\beginappendix

This appendix follows the order of the main paper.  The training data, the model, and CSA come first (Secs.~\ref{section:appendix-data}--\ref{section:appendix-csa}).  We then define the cross-view design baselines and the evaluation protocol (Secs.~\ref{section:appendix-design} and~\ref{section:appendix-evaluation}) and add generation results (Sec.~\ref{section:appendix-generation}).  The setups of the downstream experiments close the appendix (Secs.~\ref{section:appendix-worldarena}--\ref{section:appendix-real-robot}).  Image sizes in this appendix are width$\times$height, and latent and token grids list height first.

\section{Training Data and Curation}
\label{section:appendix-data}

\begin{table}[!htb]
    \centering
    \begin{minipage}{0.62\textwidth}
    \centering
    \small
    \setlength{\tabcolsep}{3pt}
    \caption{Training clips per source after holdout removal and the three-view cap.  Counts are clips, not demonstrations.}
    \label{table:pretraining-data}
    \begin{tabular*}{\linewidth}{@{\extracolsep{\fill}}lrrrr@{}}
        \toprule
        Source & 1-view & 2-view & 3-view & Total \\
        \midrule
        AgiBotWorld-Beta & 0 & 0 & 764,058 & 764,058 \\
        Galaxea & 0 & 0 & 121,735 & 121,735 \\
        RoboCOIN & 6,464 & 0 & 375,864 & 382,328 \\
        RoboMIND & 67,527 & 21,738 & 58,988 & 148,253 \\
        RoVid-X & 1,574,280 & 0 & 0 & 1,574,280 \\
        \midrule
        Total & 1,648,271 & 21,738 & 1,320,645 & 2,990,654 \\
        \bottomrule
    \end{tabular*}
    \end{minipage}
\end{table}

We first describe how we build the training clips and how training draws views from them.

\parahead{Sources.}
Tab.~\ref{table:pretraining-data} lists the clips per source.  The real-robot clips come from 14 embodiments: AgiBot G1/A2D, Galaxea R1 Lite, Leju Kuavo, AgileX Cobot Magic, AlphaBot~2, Tien Kung, Realman RMC-AIDA-L, Galbot G1, Unitree G1, single-arm Franka, UR5e, AIRBOT MMK2, dual-arm Franka FR3, and Tianqin A2.  They span single-arm, dual-arm, and humanoid robots with different camera rigs.  RoboMIND simulation and RoVid-X supply the remaining clips.

\parahead{Curation.}
Curation removes clips that would teach the model little.  A clip counts as near-static when the median adjacent-frame mean absolute luma difference of its primary view is below 0.25.  Captions are clip-level for 2,620,273 clips and episode- or task-level for 370,381, and deterministic rules normalize them and reject invalid ones.  These two filters keep 2,992,673 of 3,054,470 candidate clips.  Removing the 2,019 clips of the \benchname episodes then leaves 2,990,654.  Frames are resized with antialiasing to cover $640\times480$ and center-cropped.

\parahead{View counts and sampling.}\looseness=-1 
Tab.~\ref{table:pretraining-data} also splits the clips by view count.  All RoVid-X clips have one view, whereas most clips of the other sources have three.  VCC (Sec.~\ref{section:view-curriculum}) then decides how many of a clip's views each sample uses.  Its horizon $\tau$ is 20K steps, and each of the three view counts receives one third of the curriculum steps in expectation.  Runs without VCC sample clips uniformly and group each step by view count.

\parahead{Data-scaling subsets.}
The nested 25\% and 50\% subsets of Fig.~\ref{fig:latency-scaling}b keep whole episodes.  Stratification by data source and by the native view count of each clip keeps the full data mix in every subset.

\section{Model and Training Details}
\label{section:appendix-reproducibility}

\begin{figure*}[t]
    \centering
    \includegraphics[width=\linewidth]{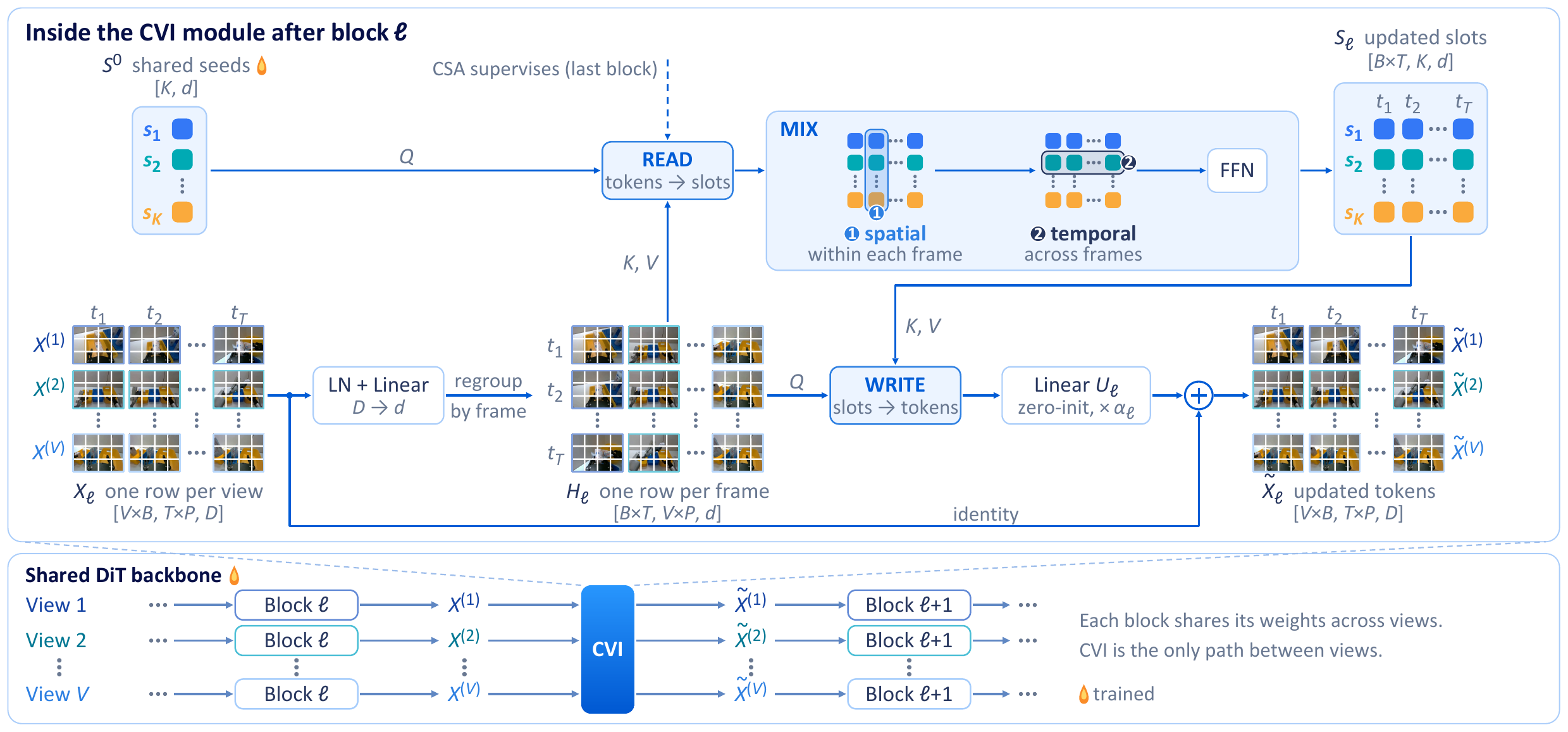}
    \caption{Data flow of one CVI module.  Bottom: each backbone block runs on every view with shared weights, and a CVI module follows every block.  Top: one CVI module (Sec.~\ref{section:shared-slot}).  $X_\ell$ has one row per view, and $H_\ell$ has one row per latent frame, holding all views.  Slot colors follow Fig.~\ref{fig:method-overview}.  $B$, $T$, $P$, $V$, and $K$ denote the batch size and the numbers of latent frames, patches per frame, views, and slots; $D=3072$ and $d=256$.}
    \label{fig:cvi-internals}
\end{figure*}

The details below complement Sec.~\ref{section:method}: the backbone, the CVI module, and the training objective.

\parahead{Backbone.}
\method builds on Wan2.2-TI2V-5B~\citep{wan2025}, a DiT with 30 blocks of width 3072 and 24 attention heads.  Its VAE maps each view to a $48\times21\times30\times40$ latent, and $2\times2$ patches give 6,300 tokens per view.  All views share the UMT5 encoding of one prompt: ``A video recorded from a robot's point of view executing the following instruction:'' followed by the instruction.  Sampling uses the default Wan negative prompt for classifier-free guidance.

\parahead{CVI modules.}
Fig.~\ref{fig:cvi-internals} traces the data flow of one module.  The module stays small because it works at width $d=256$ instead of 3072.  Its input LayerNorm has no affine parameters, and the MIX feed-forward network has hidden width 512 and uses GELU.  Every block has its own READ, MIX, and WRITE weights.  All blocks share the slot seeds $S^0$, drawn from $\mathcal{N}(0,0.02^2)$.  One module has 2.894M parameters.

\parahead{Training objective.}
The flow-matching loss averages the squared error over all latent elements of the active views except their conditioning frames.  Wan2.2-5B (FT) and the ablations start from the same backbone and use the optimizer settings of Sec.~\ref{section:experimental-setting}.

\section{Cross-View Slot Agreement Details}
\label{section:appendix-csa}

CSA relies on a teacher that links similar regions across views.  This section describes the teacher, shows why it uses VGGT-$\Omega$, and explains how the loss is computed.

\parahead{Teacher features.}
The teacher needs features that can match patches across views.  We extract them at pixel frames $0,4,\dots,80$, one per latent frame.  \mbox{DINOv2} ViT-L/14 without registers~\citep{oquab2024dinov2} encodes each view separately.  The VGGT-$\Omega$ aggregator~\citep{wang2026vggtomega} processes all views of one frame jointly, and we take its patch tokens after the global attention of the last block.  Each teacher grid is pooled onto the $15\times20$ token grid, weighting every teacher patch by its pixel overlap with each token footprint.  The pooled features are $\ell_2$-normalized.  VGGT-$\Omega$ only decides which patches correspond in Alg.~\ref{alg:teacher-relations}, and its image-position cues never enter the clustered DINOv2 features.  The spherical $K$-means step of Alg.~\ref{alg:teacher-relations} initializes its cluster centers by farthest-point sampling.

\begin{figure*}[t]
    \centering
    \includegraphics[width=\linewidth]{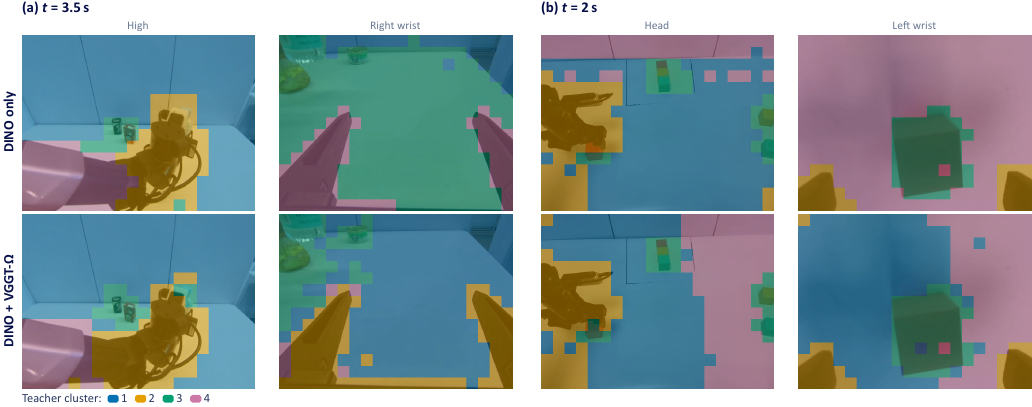}
    \caption{CSA teacher labels from DINOv2 alone (top) and with VGGT-$\Omega$ (bottom) on two \benchname episodes, each in a wrist view and a non-wrist view.  Without VGGT-$\Omega$, the table seen at close range by the wrist camera falls into a different cluster than in the non-wrist view.  With VGGT-$\Omega$, the table falls into the same cluster in both views.}
    \label{fig:teacher-dino-only}
\end{figure*}
\parahead{Why the teacher uses VGGT-$\Omega$.}\looseness=-1 
Appearance alone can put one surface into different clusters in different views.  A DINO-only teacher runs the same spherical $K$-means on the DINOv2 features of all views, without correspondences.  On a fixed sample of 131 multi-view training clips (about 0.01\%), it keeps 82.0\% of the mutual matches of Alg.~\ref{alg:teacher-relations} in one cluster.  Our teacher keeps 99.2\%.  Fig.~\ref{fig:teacher-dino-only} shows typical failures on wrist views.

\parahead{Loss computation.}\looseness=-1 
The two means of $\mathcal{L}_{\mathrm{CSA}}$ pool all labeled cross-view pairs of a microbatch.  We compute them exactly from sums over the tokens of each label and view instead of enumerating the pairs.  READ uses standard attention normalized over patches, so each slot reads a weighted mean of patch features.  CSA renormalizes this attention over slots only to obtain each patch's slot distribution; the forward pass is unchanged.

\begin{figure*}[t]
    \centering
    \includegraphics[width=\linewidth]{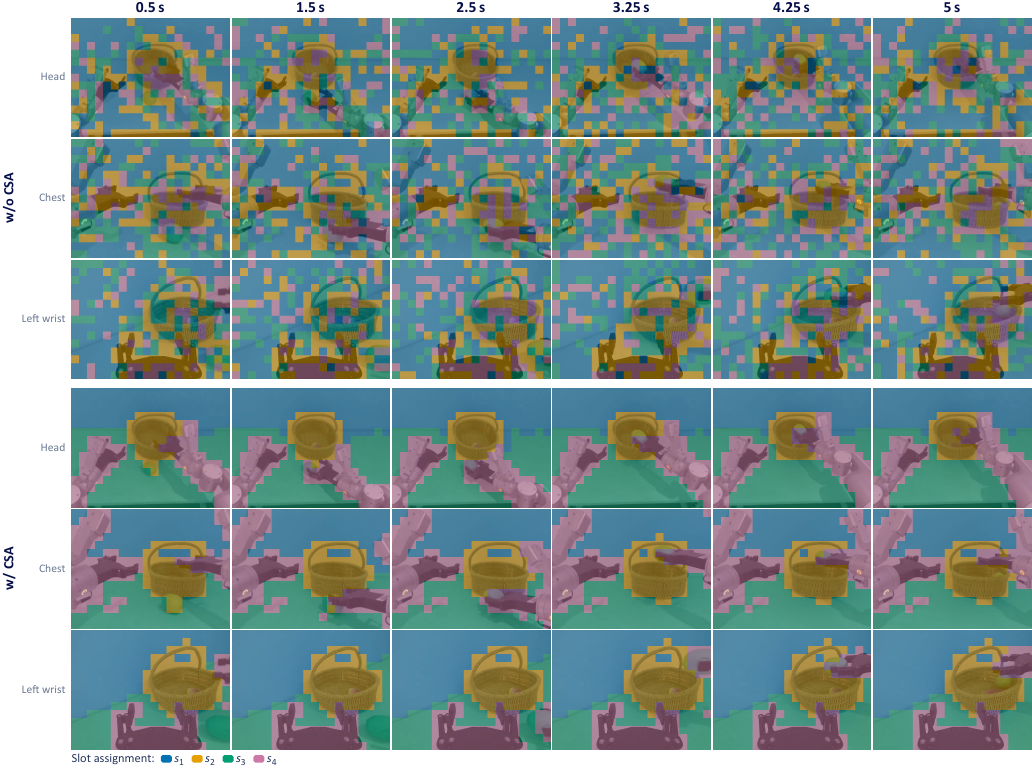}
    \caption{Slot assignments of the last READ attention over 5\,s for the episode of Fig.~\ref{fig:agreement-attention} (same colors).  The 3.25\,s column is the frame of Fig.~\ref{fig:agreement-attention}.  Without CSA (VCC+CVI), slots tend to merge similar regions, yet the grouping stays fragmented.  With CSA (\method), the basket, the arms, the table, and the background keep their slots across views and over time.}
    \label{fig:agreement-over-time}
\end{figure*}
\parahead{Slot assignments over time.}\looseness=-1 
CSA acts on each frame separately, yet its grouping persists over time.  In Fig.~\ref{fig:agreement-over-time}, the slots with CSA stay stable while the arms move.  The shared seeds and the temporal MIX likely cause this stability.

\section{Cross-View Design Baselines}
\label{section:appendix-design}

Tab.~\ref{table:design-comparison} compares CVI with two other ways to couple the views.  This section defines them and the latency measurement of Fig.~\ref{fig:latency-scaling}a.  All three designs share the data, the optimizer, and the flow-matching objective, and none uses camera embeddings.  CVI denotes the VCC+CVI model of the ablation study (Tab.~\ref{table:main-ablation}).

\begin{table}[!htb]
    \centering
    \begin{minipage}{0.62\textwidth}
    \centering
    \small
    \setlength{\tabcolsep}{1.5pt}
    \caption{DiT GPU time on one B200 (lower is better).  Per forward pass: median kernel time.  Per sample: 80 passes.  Best in bold.}
    \label{table:latency-details}
    \begin{tabular*}{\linewidth}{@{\extracolsep{\fill}}cccccccc@{}}
        \toprule
        & & \multicolumn{3}{c}{Per forward pass (ms)} & \multicolumn{3}{c}{Per sample (s)} \\
        \cmidrule(lr){3-5}\cmidrule(lr){6-8}
        Views & Tokens & Canvas & Full attn. & CVI & Canvas & Full attn. & CVI \\
        \midrule
        3 & 18,900 & \phantom{1,}659 & \phantom{1,}656 & \textbf{481} & 52.7 & 52.5 & \textbf{38.5} \\
        4 & 25,200 & 1,037 & 1,033 & \textbf{642} & 82.9 & 82.6 & \textbf{51.4} \\
        \bottomrule
    \end{tabular*}
    \end{minipage}
\end{table}

\parahead{Full attention.}\looseness=-1 
Full attention lets every token attend to all views.  Every self-attention layer runs over the concatenated tokens of all active views, and its computation thus grows quadratically with the view count.  As in \method, every view keeps its own 3D RoPE grid, and tokens at the same position in different views share their rotary phases.  The VAE still encodes each view separately with its own first-frame condition.

\begin{table*}[t]
    \centering
    \small
    \setlength{\tabcolsep}{3.5pt}
    \caption{\benchname results with one and two views.  A single view has no camera pair and thus no CVC.  Genie-Envisioner needs a scene camera and both wrist cameras and is omitted.  Parentheses give the change of \method from Wan2.2-5B (FT).  Best in bold.}
    \label{table:fewer-views}
    \begin{tabular*}{\textwidth}{@{\extracolsep{\fill}}lccccccccc@{}}
        \toprule
        \multirow{2}{*}{Method}
        & \multicolumn{4}{c}{1-view}
        & \multicolumn{5}{c}{2-view} \\
        \cmidrule(lr){2-5} \cmidrule(lr){6-10}
        & SSIM $\uparrow$ & LPIPS $\downarrow$ & FID $\downarrow$ & FVD $\downarrow$
        & SSIM $\uparrow$ & LPIPS $\downarrow$ & FID $\downarrow$ & FVD $\downarrow$ & CVC $\uparrow$ \\
        \midrule
        Wan2.2-5B              & 0.523 & 0.505 & 38.87 & 357.03 & 0.518 & 0.527 & 27.40 & 291.52 & 56.30 \\
        Wan2.1-14B             & 0.511 & 0.524 & 31.99 & 294.16 & 0.496 & 0.556 & 26.82 & 266.73 & 57.39 \\
        Wan2.2-A14B            & 0.534 & 0.490 & 27.28 & 245.34 & 0.508 & 0.539 & 22.46 & 233.82 & 56.56 \\
        Cosmos3-Edge & 0.617 & 0.422 & 44.01 & 406.40 & 0.578 & 0.517 & 42.79 & 377.79 & 45.70 \\
        Cosmos3-Nano & 0.677 & 0.332 & 22.06 & 104.23 & 0.666 & 0.370 & 19.54 & 100.75 & 75.17 \\
        \midrule
        Wan2.2-5B (FT)         & 0.702 & 0.292 & 16.46 & 63.93  & 0.675 & 0.342 & 12.24 & 39.55  & 69.97 \\
        \textsf{\bfseries Ours} & \improve{\textbf{0.705}}{+0.003} & \improve{\textbf{0.283}}{$-0.009$} & \improve{\textbf{15.78}}{$-0.68$} & \improve{\textbf{60.23}}{$-3.70$} & \improve{\textbf{0.693}}{+0.018} & \improve{\textbf{0.318}}{$-0.024$} & \improve{\textbf{10.89}}{$-1.35$} & \improve{\textbf{34.03}}{$-5.52$} & \improve{\textbf{75.93}}{+5.96} \\
        \bottomrule
    \end{tabular*}
\end{table*}

\begin{table*}[t]
    \centering
    \small
    \setlength{\tabcolsep}{3.5pt}
    \caption{\benchname fidelity by camera type.  Wrist views have ``wrist'' or ``hand'' in the camera name.  The three-view setting of Genie-Envisioner uses two wrist views per episode.  Best in bold.}
    \label{table:camera-type}
    \begin{tabular*}{\textwidth}{@{\extracolsep{\fill}}lcccccccc@{}}
        \toprule
        \multirow{3}{*}{Method}
        & \multicolumn{4}{c}{3-view} & \multicolumn{4}{c}{4-view} \\
        \cmidrule(lr){2-5}\cmidrule(lr){6-9}
        & \multicolumn{2}{c}{SSIM $\uparrow$} & \multicolumn{2}{c}{LPIPS $\downarrow$} & \multicolumn{2}{c}{SSIM $\uparrow$} & \multicolumn{2}{c}{LPIPS $\downarrow$} \\
        \cmidrule(lr){2-3}\cmidrule(lr){4-5}\cmidrule(lr){6-7}\cmidrule(lr){8-9}
        & Wrist & Other & Wrist & Other & Wrist & Other & Wrist & Other \\
        \midrule
        Wan2.2-5B & 0.497 & 0.521 & 0.586 & 0.497 & 0.497 & 0.521 & 0.583 & 0.497 \\
        Wan2.1-14B & 0.458 & 0.506 & 0.625 & 0.526 & 0.458 & 0.508 & 0.624 & 0.521 \\
        Wan2.2-A14B & 0.448 & 0.533 & 0.647 & 0.487 & 0.447 & 0.536 & 0.647 & 0.485 \\
        Genie-Envisioner & \textbf{0.684} & 0.692 & 0.436 & 0.388 & \textbf{0.691} & 0.622 & 0.430 & 0.474 \\
        Cosmos3-Edge & 0.587 & 0.593 & 0.557 & 0.509 & 0.651 & 0.641 & 0.478 & 0.442 \\
        Cosmos3-Nano & 0.605 & 0.660 & 0.468 & 0.397 & 0.670 & 0.680 & 0.404 & 0.354 \\
        \midrule
        Wan2.2-5B (FT) & 0.624 & 0.692 & 0.439 & 0.301 & 0.627 & 0.689 & 0.435 & 0.306 \\
        \textsf{\bfseries Ours} & 0.664 & \textbf{0.706} & \textbf{0.389} & \textbf{0.282} & 0.669 & \textbf{0.708} & \textbf{0.384} & \textbf{0.279} \\
        \bottomrule
    \end{tabular*}
\end{table*}

\parahead{Pixel canvas.}
The canvas baseline instead turns all views into one wide video.  The views are tiled left to right in sampler order into one $640V\times480$ video, which the frozen VAE encodes once.  The first canvas frame is the image condition, and evaluation cuts the generated canvas back into $640\times480$ views.  Each tile yields $15\times20$ tokens per latent frame.  The canvas thus holds as many tokens as full attention over all views.

\parahead{Latency.}
\looseness=-1 Tab.~\ref{table:latency-details} and Fig.~\ref{fig:latency-scaling}a report DiT GPU time per sample.  A sample takes 80 forward passes: 40 Euler steps with two guidance branches each.  The time per forward pass is the median of five repeats and excludes the text encoder.  For three to eight views, one sample takes 38.5 to 104\,s with CVI and 52.5 to 263\,s with full attention.  Canvas requires nearly as much computation as full attention, as its equal token count suggests.

\begin{figure*}[t]
    \centering
    \includegraphics[width=\linewidth]{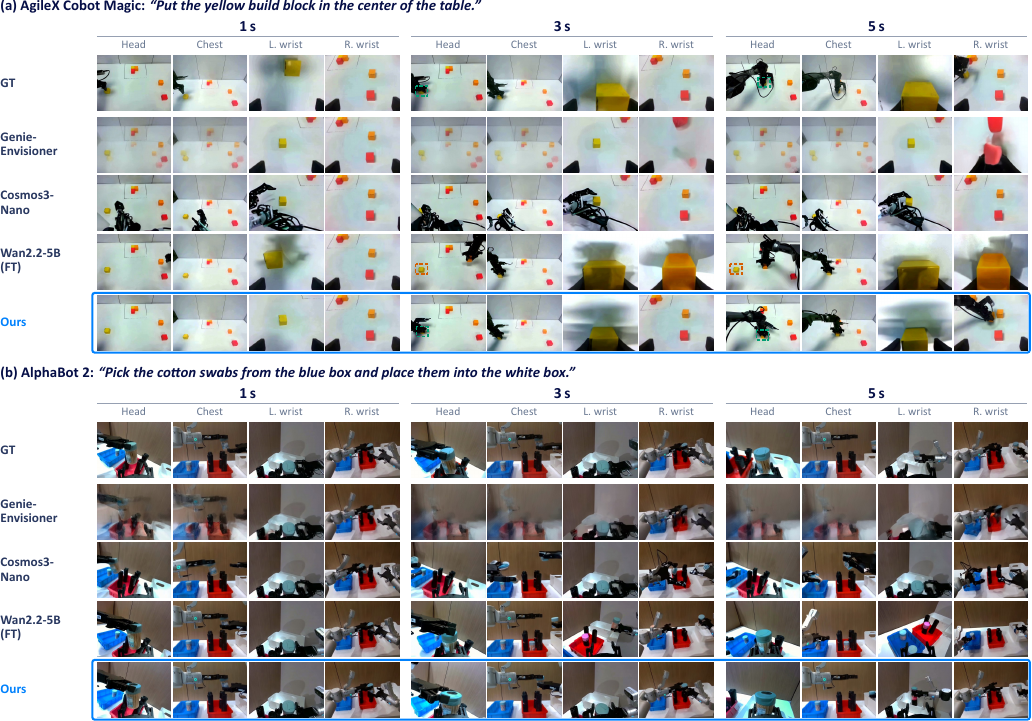}
    \caption{Four-view \benchname episodes in the layout of Fig.~\ref{fig:qualitative-results}.  Green boxes mark the block in the left gripper, and orange boxes mark errors of Wan2.2-5B (FT).  In (a), its head view leaves the yellow block on the table, while its chest and left wrist views show it grasped.}
    \label{fig:qualitative-more-4view}
\end{figure*}
\section{Evaluation Protocol}
\label{section:appendix-evaluation}

The \benchname numbers of Tab.~\ref{table:mvwm-bench} depend on a few protocol choices, which this section fixes.

\parahead{View selection.}
The three-view setting keeps the first listed view and picks two of the other views by a fixed pseudo-random ranking.  Genie-Envisioner instead needs a scene camera and both wrist cameras, so its three views are these cameras.  The four-view setting uses all four views.  It tests an unseen view count and new camera placements.  The ground truth uses the resize and center crop of training.

\begin{figure*}[p]
    \centering
    \includegraphics[height=0.93\textheight]{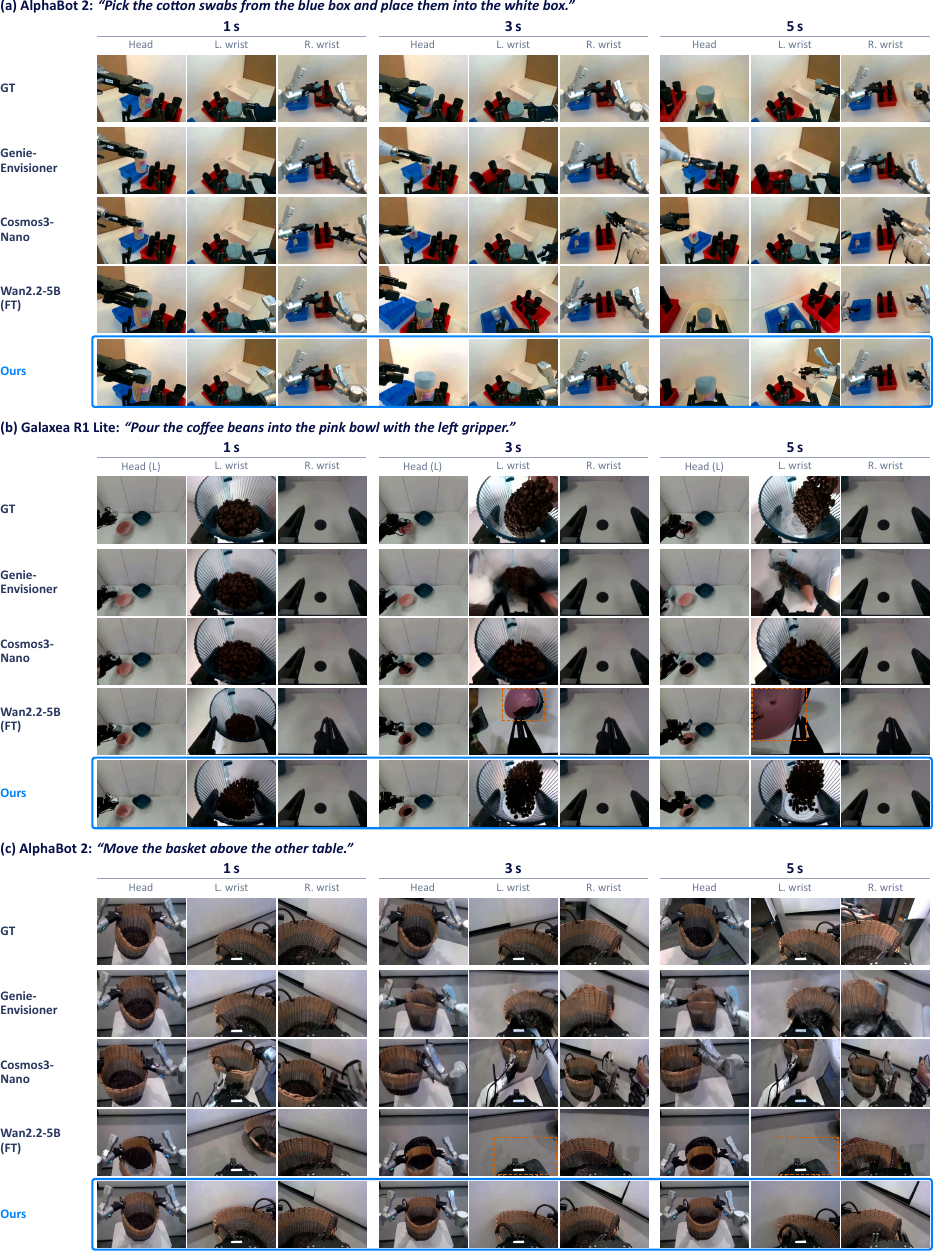}
    \caption{Three-view \benchname episodes in the layout of Fig.~\ref{fig:qualitative-results}.  Orange boxes mark errors of Wan2.2-5B (FT), and Head (L) is the left head camera.  In (b), it replaces the glass bowl of coffee beans in the left wrist view with a pink bowl.  In (c), its left wrist view loses the basket, although its head view still shows it.}
    \label{fig:qualitative-more-3view}
\end{figure*}
\parahead{Fidelity metrics.}
\looseness=-1 SSIM and LPIPS~\citep{zhang2018lpips} with AlexNet features compare each generated frame with its ground truth.  We average them over the frames of a view, then over the views of an episode, and then over episodes.  FID and FVD instead compare distributions.  FID~\citep{heusel2017fid} uses the 2048-D Inception pool features of pytorch-fid on every frame, resized to $299\times299$.  FVD uses the StyleGAN-V~\citep{skorokhodov2022styleganv} implementation of I3D~\citep{carreira2017i3d} on non-overlapping 16-frame clips of each view and drops the incomplete tail.  Both pool all views and episodes into one dataset-level statistic, and views are never tiled into one image.  The radar in Fig.~\ref{fig:teaser} plots the three-view results of Tab.~\ref{table:mvwm-bench} on linear axes, and lower LPIPS, FID, and FVD lie farther out.  From center to rim, the axes span CVC 52 to 80, FVD 110 to 10, FID 24 to 6, LPIPS 0.50 to 0.30, and SSIM 0.60 to 0.70, so every value lies inside.

\parahead{CVC.}
CVC builds on the official MEt3R~\citep{asim2025met3r} code.  MEt3R splats FeatUp-upsampled DINO ViT-S/16 features of both frames into the first camera with MASt3R point maps.  The dissimilarity is the mean cosine distance over the region that both frames cover, and the overlap is the fraction of the image this region fills.  Frames are resized without cropping, and each unordered camera pair is scored once in camera order.  A pair without an eligible frame is left out.

\section{Additional Generation Results}
\label{section:appendix-generation}

We extend the generation results of Secs.~\ref{section:mv-generation} and~\ref{section:ac-generation} with fewer views, more qualitative examples, camera types, a reference-free VLM judge, and action-conditioned rollouts.

\begin{figure*}[t]
    \centering
    \includegraphics[width=\linewidth]{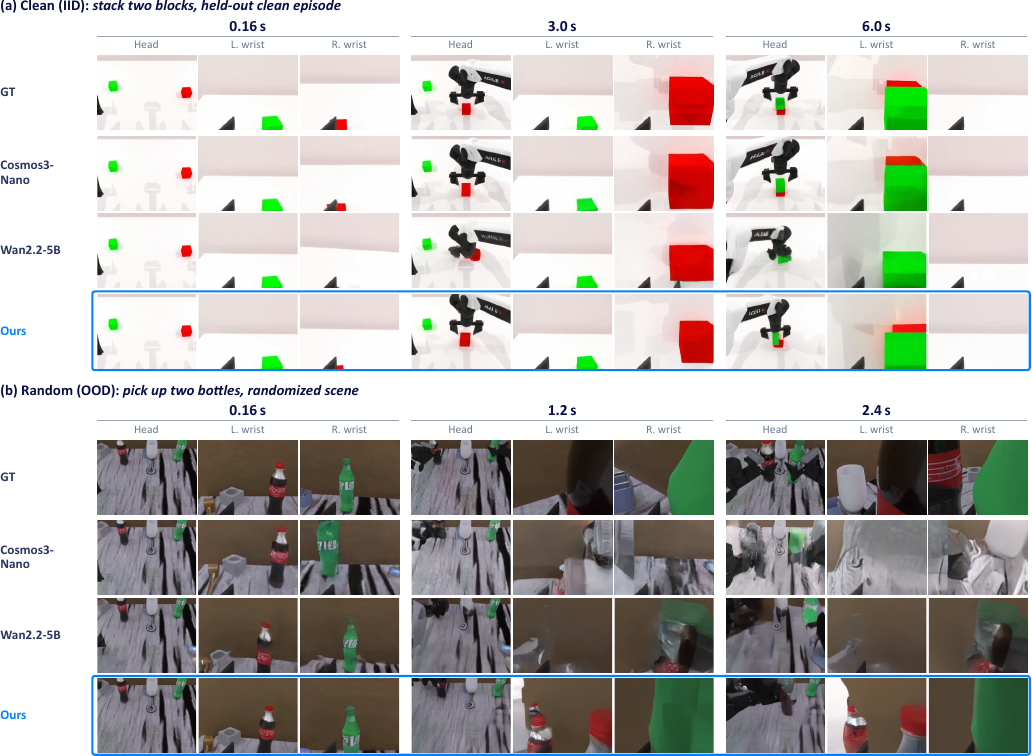}
    \caption{Action-conditioned video generation on RoboTwin 2.0 after four epochs of fine-tuning.  (a)~A held-out Clean episode (stack two blocks) at 0.16, 3.0, and 6.0\,s.  (b)~A Random episode (pick up two bottles) at 0.16, 1.2, and 2.4\,s.  All models start from the real first frame of each view and follow the same recorded trajectory.  Under randomization, Wan2.2-5B blurs, Cosmos3-Nano washes out the wrist views, and our AC-WM keeps the wrist views sharp.  Episodes are selected for illustration.}
    \label{fig:worldarena-qualitative}
\end{figure*}
\parahead{Fewer views.}
Tab.~\ref{table:fewer-views} tests whether the gains hold with one or two views.  Both settings use the same 500 episodes, and their cameras are a subset of the three-view cameras.  The single view is always a body-mounted or external camera.  Two-view CVC covers the 484 episodes whose camera pair reaches the overlap threshold.  \method achieves the best value of every metric with both view counts.  With a single view, \method still improves slightly over Wan2.2-5B (FT).  Multi-view training thus does not seem to hurt single-view generation.  The two-view CVC gain over Wan2.2-5B (FT) is 6.0 points, close to the three- and four-view gains (6.4 and 5.2).

\parahead{More examples.}
Figs.~\ref{fig:qualitative-more-4view} and~\ref{fig:qualitative-more-3view} add five episodes from three embodiments in the layout of Fig.~\ref{fig:qualitative-results}.  \method again follows the ground-truth motion and keeps the views consistent, whereas the wrist and head views of Wan2.2-5B (FT) often disagree with each other.

\parahead{Camera types.}
Wrist views are the hardest views for most generators (Tab.~\ref{table:camera-type}).  With three views, every method has lower SSIM and higher LPIPS on wrist views than on the other views.  With four views, Genie-Envisioner scores better on wrist views in both SSIM and LPIPS, and Cosmos3-Edge reaches a higher wrist SSIM.  \method has the lowest LPIPS on both camera types in both settings.

\begin{table*}[t]
    \centering
    \small
    \setlength{\tabcolsep}{3.5pt}
    \caption{VLM judge scores on \benchname with three views (1--5, higher is better).  Overall is the mean of the four criteria, and real videos give a reference.  Parentheses give the change of \method from Wan2.2-5B (FT).  Best generated value in bold.}
    \label{table:vlm-judge}
    \begin{tabular*}{\textwidth}{@{\extracolsep{\fill}}lccccc@{}}
        \toprule
        Method & \shortstack{Instruction\\following $\uparrow$} & \shortstack{First-frame\\grounding $\uparrow$} & \shortstack{Physical\\plausibility $\uparrow$} & \shortstack{Cross-view\\consistency $\uparrow$} & Overall $\uparrow$ \\
        \midrule
        Wan2.2-5B & 2.126 & 2.120 & 2.000 & 2.010 & 2.064 \\
        Wan2.1-14B & 2.300 & 2.152 & 2.012 & 2.000 & 2.116 \\
        Wan2.2-A14B & 2.642 & 2.206 & 2.024 & 2.002 & 2.219 \\
        Genie-Envisioner & 2.068 & 2.966 & 2.216 & 2.394 & 2.411 \\
        Cosmos3-Edge & 2.052 & 2.254 & 2.044 & 2.046 & 2.099 \\
        Cosmos3-Nano & 2.338 & 3.070 & 2.504 & 2.392 & 2.576 \\
        \midrule
        Wan2.2-5B (FT) & 2.942 & 2.944 & 2.434 & 2.218 & 2.635 \\
        \textsf{\bfseries Ours} & \improve{\textbf{2.976}}{+0.034} & \improve{\textbf{3.154}}{+0.210} & \improve{\textbf{2.648}}{+0.214} & \improve{\textbf{2.474}}{+0.256} & \improve{\textbf{2.813}}{+0.178} \\
        \midrule
        \textcolor{gray}{Real videos} & \textcolor{gray}{3.226} & \textcolor{gray}{3.824} & \textcolor{gray}{3.208} & \textcolor{gray}{3.418} & \textcolor{gray}{3.419} \\
        \bottomrule
    \end{tabular*}
\end{table*}

\begin{table*}[t]
    \centering
    \small
    \setlength{\tabcolsep}{3.5pt}
    \caption{\looseness=-1 Action-conditioned world models on RoboTwin 2.0.  All four receive three camera views and numeric controls but no text, train for four epochs at a batch size of 512 and a base learning rate of $4\times10^{-6}$, and sample 50 steps with shift 5 and no guidance.  All parameters include frozen modules such as the VAE, which trainable parameters exclude.}
    \label{table:ac-settings}
    \begin{tabular*}{\textwidth}{@{\extracolsep{\fill}}lcccc@{}}
        \toprule
        Setting & Cosmos3-Edge & Cosmos3-Nano & Wan2.2-5B & \textsf{\bfseries Ours} \\
        \midrule
        Initialization & \multicolumn{2}{c}{official checkpoints} & Wan2.2-TI2V-5B & \method \\
        All parameters & 4.07B & 15.88B & 5.76B & 5.84B \\
        Trainable parameters & 1.42B & 6.97B & 5.05B & 5.14B \\
        Added parameters & \multicolumn{2}{c}{none} & \multicolumn{2}{c}{51.1M (EVAC)} \\
        Views & \multicolumn{2}{c}{one head-over-wrists mosaic} & \multicolumn{2}{c}{three per-view latents} \\
        Head / wrist size (W$\times$H) & \multicolumn{2}{c}{$320\times256$ / $160\times128$} & \multicolumn{2}{c}{$320\times256$ / $320\times256$} \\
        Cross-view exchange & \multicolumn{2}{c}{attention within the mosaic} & joint self-attention & CVI (86.8M) \\
        \midrule
        Actions per clip & \multicolumn{2}{c}{16 joint targets (14-D)} & \multicolumn{2}{c}{16 end-effector deltas (14-D)} \\
        Action maps & \multicolumn{2}{c}{none} & \multicolumn{2}{c}{added to the patch tokens} \\
        Action tokens & \multicolumn{2}{c}{16, in self-attention} & \multicolumn{2}{c}{16 from a resampler, in cross-attention} \\
        State token & \multicolumn{2}{c}{none} & \multicolumn{2}{c}{1, in cross-attention} \\
        \midrule
        Optimizer & \multicolumn{2}{c}{AdamW $(0.9, 0.99)$, weight decay 0.05} & \multicolumn{2}{c}{AdamW $(0.9, 0.95)$, weight decay 0.01} \\
        Sampler & \multicolumn{2}{c}{UniPC~\citep{zhao2023unipc}} & \multicolumn{2}{c}{Euler} \\
        \bottomrule
    \end{tabular*}
\end{table*}

\parahead{VLM judge.}
SSIM and LPIPS need the ground-truth video, and none of the metrics above checks physical plausibility directly.  We therefore also score generations with a VLM judge that needs no reference.  Gemini 4 Argon sees one composite image per episode.  Its rows are the views, and its columns are frames 0, 16, 32, 48, 64, and 80, where the first column is the given frame.  We judge all 500 three-view \benchname episodes of every generator and of the real videos, which serve as a reference (Tab.~\ref{table:vlm-judge}).

The judge rates four criteria from 1 to 5 against a fixed rubric.  Instruction following asks whether the robot visibly performs, or clearly progresses on, the instructed task.  First-frame grounding asks whether the rollout keeps the scene, objects, and robot of the given frame.  Physical plausibility asks whether the robot and the objects keep their identity and shape, move only after contact, and never pass through each other.  Cross-view consistency asks whether all views of a column show the same object positions, gripper state, and task stage.  The overall score is the mean of the four.

\method scores highest among the generators on every criterion and overall.  Its gains over Wan2.2-5B (FT) are 0.21 in first-frame grounding, 0.21 in physical plausibility, 0.26 in cross-view consistency, and 0.18 overall.  All four 95\% paired bootstrap intervals exclude zero.  The zero-shot Wan models generate each view alone and receive a cross-view score of 2 in nearly every episode.  Real videos score 3.42 overall and stay well above every generator.

\parahead{Action-conditioned rollouts.}
Fig.~\ref{fig:worldarena-qualitative} illustrates the randomization gap of Tab.~\ref{table:worldarena}.  All three models follow both arms in all three views of the Clean episode, and the moving wrist views of our AC-WM stay close to the ground truth.  Cosmos3-Nano washes out both wrist views of the Random episode within about one second.  For display, we upsample the $160\times128$ wrist views of Cosmos3-Nano.

\section{Action-Conditioned Generation Details}
\label{section:appendix-worldarena}

Sec.~\ref{section:ac-generation} fine-tunes each video model into an AC-WM.  The paragraphs below describe this conversion, and Tab.~\ref{table:ac-settings} summarizes the four models.

\parahead{Data.}
Each RoboTwin~2.0 task~\citep{chen2025robotwin2} is recorded by a head camera and two wrist cameras.  Training clips have 17 frames at 25\,FPS: the current frame and 16 future frames, and each future frame is paired with one recorded action.

\parahead{Action conditioning.}
Wan2.2-5B and our AC-WM share the EVAC-style conditioning~\citep{jiang2025evac} and differ only in the video backbone.  The current frame of each view serves as a clean condition.  Action maps render the end-effector poses into every camera, and the VAE encodes them.  These maps are the only input that uses camera calibration.  The networks receive no camera parameters, and, unlike EVAC, we use no ray maps.  A Perceiver resampler~\citep{jaegle2021perceiver} encodes the 16 end-effector deltas as tokens that replace the text tokens in cross-attention, together with one token for the robot state.

\parahead{Training and rollout.}
Wan2.2-5B and our AC-WM update the whole DiT and the conditioning modules.  The Cosmos3 models instead follow the official forward-dynamics recipe and update their generation tower, which processes the video and action tokens.  Their equally large understanding tower processes only text and stays frozen, as do all VAEs.  At test time, each model rolls out the whole episode in 17-frame chunks.  Each chunk starts from the last generated frame, without resetting frames to the ground truth.  At the start of each chunk, Wan2.2-5B and our AC-WM also receive the robot's joint state from the recorded episode, the same source as the actions; the Cosmos3 models instead receive absolute joint targets as actions.

\section{World-Action Model Details}
\label{section:appendix-wam}

Our WAM adds CVI to FastWAM-Joint~\citep{yuan2026fastwam}.  This section describes the architecture of the policy, its initialization, and its settings on each simulation benchmark.

\begin{figure*}[t]
    \centering
    \includegraphics[width=\linewidth]{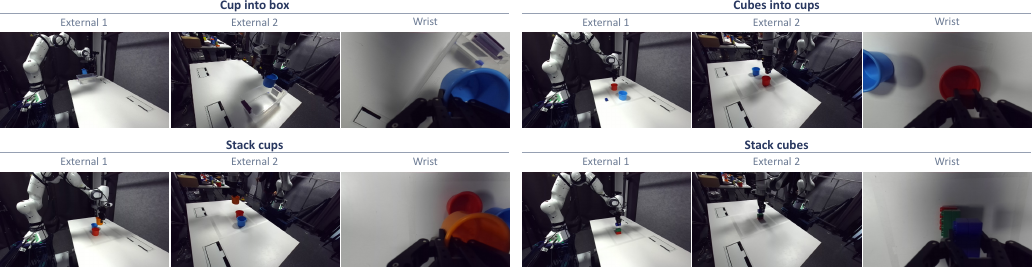}
    \caption{The four real-robot tasks in the three camera views, each shown at one instant of a teleoperated demonstration.  Cup into box was recorded only at $320\times180$ and is upsampled for display.}
    \label{fig:real-robot-tasks}
\end{figure*}
\parahead{Architecture.}
As in FastWAM-Joint, the video and action networks form an asymmetric mixture of transformers.  CVI follows every block of the video network.  The action queries of every layer attend to the video keys and values of all views after the CVI module.  Policy training uses no CSA, and the policy adds no other fusion parameters.  FastWAM and FastWAM-Joint keep their original design, tiling all views into one frame.

\parahead{Initialization.}
The video network loads every DiT tensor of \method, including the CVI modules.  The action network mirrors the Wan blocks at width 1024.  In simulation, its blocks start from the DiT weights of \method, linearly interpolated to width 1024, just as FastWAM-Joint derives its action network from Wan2.2-TI2V-5B.  Its encoder and head start from random weights.  On the real robot, it keeps the official FastWAM-Joint initialization.

\begin{table}[!htb]
    \centering
    \begin{minipage}{0.62\textwidth}
    \centering
    \small
    \setlength{\tabcolsep}{3pt}
    \caption{Hyperparameters of our WAM on LIBERO.  LIBERO-Plus evaluation uses the same inference settings.}
    \label{table:hparams-libero}
    \begin{tabular*}{\linewidth}{@{\extracolsep{\fill}}ll@{}}
        \toprule
        Hyperparameter & Value \\
        \midrule
        Demonstrations (tasks) & 1,712 (40) \\
        Views (W$\times$H); frames per view & 2 $\times$ 224$\times$224; 9 (stride 4) \\
        Action space; chunk length & 7-D end-effector delta; 32 \\
        \midrule
        Global batch; epochs (steps) & 128; 10 (21,700) \\
        Learning rate; schedule & $10^{-4}$; 5\% warmup, cosine \\
        Optimizer ($\beta_1, \beta_2$); weight decay & AdamW (0.9, 0.95); 0.01 \\
        Gradient clipping & 1.0 \\
        Augmentation & None \\
        Loss weights (video / action) & 1 / 1 \\
        \midrule
        Flow shift, video / action (train; test) & 5 / 1; 5 / 5 \\
        Denoising steps; executed actions & 10; 10 of 32 \\
        \bottomrule
    \end{tabular*}
    \end{minipage}
\end{table}

\parahead{LIBERO and LIBERO-Plus.}
Tab.~\ref{table:hparams-libero} lists the LIBERO settings.  The 10,030 LIBERO-Plus instances comprise 1,599 Camera, 1,550 Robot, 1,537 Language, 1,142 Light, 1,076 Background, 1,601 Noise, and 1,525 Layout instances.  The total success rate divides all successes by 10,030 rather than averaging categories.  In Tabs.~\ref{table:policy-model-comparison} and~\ref{table:bimanual-policy-comparison}, the rows other than FastWAM-Joint and ours are published results~\citep{wang2026openwam,chen2025robotwin2,chen2026robodojo}, so their totals follow the sources, and Avg is computed before rounding.  We reproduce FastWAM-Joint with its official code, except for its published LIBERO result.  The ablation checkpoints in Tab.~\ref{table:main-ablation} use the same full benchmark.

\begin{figure}[!htb]
    \centering
    \begin{minipage}{0.62\textwidth}
    \centering
    \includegraphics[width=\linewidth]{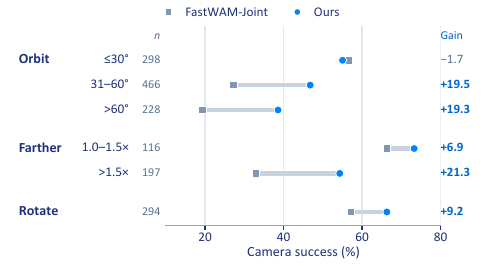}
    \caption{LIBERO-Plus Camera success by perturbation ($n$ per row).  The policies see only the default cameras in training.  At test time, the third-person camera orbits the scene (binned by the larger of the azimuth and elevation offsets), moves away, or rotates in place.  Gain: ours minus FastWAM-Joint.}
    \label{fig:libero-plus-camera}
    \end{minipage}
\end{figure}

\parahead{Camera viewpoints.}
LIBERO-Plus moves only the third-person camera, and the wrist camera stays fixed.  Fig.~\ref{fig:libero-plus-camera} splits the 1,599 Camera instances by the type and size of the change.  The gain of our WAM over FastWAM-Joint grows with the shift.  The two policies are on par for orbits up to 30$^\circ$.  Beyond 30$^\circ$, our WAM raises success from 24.6\% to 44.1\%.  It also gains 21.3 points when the camera is more than 1.5 times farther away and 9.2 points when the camera only rotates in place.

\begin{table}[t]
    \centering
    \begin{minipage}{0.62\textwidth}
    \centering
    \small
    \setlength{\tabcolsep}{3pt}
    \caption{Separating CVI from the \method prior (success, \%).  Wan + CVI has our architecture and recipe but starts from Wan2.2-TI2V-5B, like FastWAM-Joint.  Best in bold.}
    \label{table:wam-init}
    \begin{tabular*}{\linewidth}{@{\extracolsep{\fill}}lccc@{}}
        \toprule
        & FastWAM-Joint & Wan + CVI & \textsf{\bfseries Ours} \\
        \midrule
        Initialization & Wan2.2-5B & Wan2.2-5B & \method \\
        CVI & & $\checkmark$ & $\checkmark$ \\
        \midrule
        LIBERO & 98.5 & \textbf{98.7} & 98.6 \\
        LIBERO-Plus & 70.0 & 70.1 & \textbf{74.9} \\
        \quad Camera & 40.7 & 46.0 & \textbf{53.6} \\
        \quad Robot & \textbf{71.5} & 61.8 & \textbf{71.5} \\
        \quad Language & \textbf{94.7} & 87.2 & 93.3 \\
        \quad Light & \textbf{96.3} & 89.6 & 94.7 \\
        \quad Background & 58.0 & 62.3 & \textbf{63.6} \\
        \quad Noise & 55.3 & \textbf{69.4} & 69.1 \\
        \quad Layout & 78.8 & 78.2 & \textbf{81.0} \\
        \bottomrule
    \end{tabular*}
    \end{minipage}
\end{table}

\parahead{Video initialization.}\looseness=-1 
Our WAM differs from FastWAM-Joint in two ways: it adds CVI, and it starts from \method.  To separate the two, we train Wan + CVI with our architecture and recipe but the initialization of FastWAM-Joint.  Both of its networks start from Wan2.2-TI2V-5B, and its CVI modules start as at the beginning of pretraining, with zero WRITE projections.  CVI alone leaves LIBERO-Plus success nearly unchanged (70.1\% vs.\ 70.0\%) yet raises Camera success from 40.7\% to 46.0\% (Tab.~\ref{table:wam-init}).  Starting from \method adds 4.8 points overall and 7.6 points on Camera.  The multi-view prior thus provides most of the policy gain.

\begin{table}[tbp]
    \centering
    \begin{minipage}{0.62\textwidth}
    \centering
    \small
    \setlength{\tabcolsep}{3pt}
    \caption{Hyperparameters of our WAM for Clean2Random on RoboTwin 2.0.  ColorJitter strengths are 0.2, except 0.05 for hue.}
    \label{table:hparams-robotwin}
    \begin{tabular*}{\linewidth}{@{\extracolsep{\fill}}ll@{}}
        \toprule
        Hyperparameter & Value \\
        \midrule
        Demonstrations (tasks) & 2,500 clean (50) \\
        Views (W$\times$H); frames per view & 3 $\times$ 320$\times$256; 9 (stride 4) \\
        Action space; chunk length & 14-D joint position; 32 \\
        \midrule
        Global batch; epochs (steps) & 1,024; 5 (2,685) \\
        Learning rate; schedule & $10^{-4}$; 5\% warmup, cosine \\
        Optimizer ($\beta_1, \beta_2$); weight decay & AdamW (0.9, 0.95); 0.01 \\
        Gradient clipping & 1.0 \\
        Augmentation & ColorJitter, $p=0.8$ \\
        Loss weights (video / action) & 1 / 1 \\
        \midrule
        Flow shift, video / action (train; test) & 5 / 1; 5 / 5 \\
        Denoising steps; executed actions & 10; 24 of 32 \\
        \bottomrule
    \end{tabular*}
    \end{minipage}
\end{table}

\begin{table}[tbp]
    \centering
    \begin{minipage}{0.62\textwidth}
    \centering
    \small
    \setlength{\tabcolsep}{3pt}
    \caption{Hyperparameters of our WAM on RoboDojo.  One percent of the 3,500 demonstrations is held out for validation.}
    \label{table:hparams-robodojo}
    \begin{tabular*}{\linewidth}{@{\extracolsep{\fill}}ll@{}}
        \toprule
        Hyperparameter & Value \\
        \midrule
        Demonstrations (tasks) & 3,465 (35) \\
        Views (W$\times$H); frames per view & 3 $\times$ 320$\times$256; 9 (stride 4) \\
        Action space; chunk length & 14-D joint position; 32 \\
        \midrule
        Global batch; epochs (steps) & 256; 16 (114,704) \\
        Learning rate; schedule & $10^{-4}$; 5\% warmup, cosine \\
        Optimizer ($\beta_1, \beta_2$); weight decay & AdamW (0.9, 0.95); 0.01 \\
        Gradient clipping & 1.0 \\
        Augmentation & None \\
        Loss weights (video / action) & 1 / 1 \\
        \midrule
        Flow shift, video / action (train; test) & 5 / 1; 5 / 5 \\
        Denoising steps; executed actions & 10; 24 of 32 \\
        \bottomrule
    \end{tabular*}
    \end{minipage}
\end{table}

\begin{table*}[t]
    \centering
    \small
    \setlength{\tabcolsep}{4pt}
    \caption{Real-world task suite and success criteria.  Every policy receives the same instruction.  A trial succeeds when its success criterion holds within 60\,s.  The score of a two-step task counts a trial that completes one of its two subtasks as half a success.}
    \label{table:real-robot-tasks}
    \begin{tabular*}{\textwidth}{@{\extracolsep{\fill}}l>{\raggedright\arraybackslash}p{0.31\textwidth}>{\raggedright\arraybackslash}p{0.50\textwidth}@{}}
        \toprule
        Task & Instruction & Success criterion \\
        \midrule
        Cup into box & Put the blue cup in the box. & The blue cup comes to rest inside the box. \\
        \addlinespace[3pt]
        Cubes into cups & Put the red cube in the red cup and the blue cube in the blue cup. & Each cube rests inside the cup of its color, placed in either order.\newline \textit{Subtasks:} red cube into red cup; blue cube into blue cup. \\
        \addlinespace[3pt]
        Stack cups & Stack the orange cup in the red cup, then the red cup in the blue cup. & All three cups end in one nested stack, with the orange cup inside the red cup inside the blue cup.\newline \textit{Subtasks:} orange cup into red cup; then red cup into blue cup. \\
        \addlinespace[3pt]
        Stack cubes & Stack the green block on the red block, then the blue block on top. & The three cubes stand as one stack, with red at the bottom, green in the middle, and blue on top.\newline \textit{Subtasks:} green cube onto red cube; then blue cube onto green cube. \\
        \bottomrule
    \end{tabular*}
\end{table*}

\parahead{RoboTwin 2.0 and RoboDojo.}
Tabs.~\ref{table:hparams-robotwin} and~\ref{table:hparams-robodojo} list the Clean2Random and RoboDojo settings.  Both benchmarks use their official protocols and task suites.

\section{Real-Robot Setup}
\label{section:appendix-real-robot}

The real-robot experiment of Sec.~\ref{section:policy-transfer} tests whether the gains carry over to a physical robot.  This section describes its hardware, demonstrations, policies, and trial protocol.

\parahead{Hardware.}
We use a Franka Research 3 arm with a Robotiq 2F-85 gripper in the DROID setup~\citep{khazatsky2024droid}.  Two ZED 2i cameras view the workspace from the left and the right, and a ZED Mini is mounted on the wrist.  The policies use all three streams at 15\,Hz and $320\times180$ pixels.

\parahead{Tasks and demonstrations.}
Tab.~\ref{table:real-robot-tasks} lists the four tasks, and Fig.~\ref{fig:real-robot-tasks} shows them in all three views.  They range from one pick-and-place to two-step stacking.  All demonstrations are teleoperated with a Meta Quest controller.  For each task, we keep the 50 recordings closest to the median length, which gives 200 demonstrations and 58,794 frames (about 65 minutes).  Median lengths are 14.2\,s for cup into box, 19.5\,s for cubes into cups, 19.8\,s for stack cups, and 23.0\,s for stack cubes.  The two stacking tasks start from randomized positions of the objects.

\begin{table}[!htb]
    \centering
    \begin{minipage}{0.62\textwidth}
    \centering
    \small
    \setlength{\tabcolsep}{3pt}
    \caption{Hyperparameters of our WAM on the real robot.  Only its video network starts from \method.}
    \label{table:hparams-realrobot}
    \begin{tabular*}{\linewidth}{@{\extracolsep{\fill}}ll@{}}
        \toprule
        Hyperparameter & Value \\
        \midrule
        Demonstrations (tasks) & 200 (4) \\
        Views (W$\times$H); frames per view & 3 $\times$ 320$\times$192; 33 (stride 1) \\
        Action space; chunk length & 8-D joint position; 32 \\
        \midrule
        Global batch; epochs (steps) & 128; 10 (4,600) \\
        Learning rate; schedule & $10^{-4}$; 5\% warmup, cosine \\
        Optimizer ($\beta_1, \beta_2$); weight decay & AdamW (0.9, 0.95); 0.01 \\
        Gradient clipping & 1.0 \\
        Augmentation & None \\
        Loss weights (video / action) & 1 / 1 \\
        \midrule
        Flow shift, video / action (train; test) & 5 / 1; 5 / 1 \\
        Denoising steps; executed actions & 10; 24 of 32 (15\,Hz) \\
        \bottomrule
    \end{tabular*}
    \end{minipage}
\end{table}

\parahead{Policies.}\looseness=-1 
All three policies train on the same 200 demonstrations with the settings of Tab.~\ref{table:hparams-realrobot}.  They see the same three views, resized to $320\times192$.  FastWAM and FastWAM-Joint receive them tiled side by side into one $960\times192$ frame, whereas our WAM keeps them as separate views.  Each policy executes 24 of its 32 predicted joint-position actions at 15\,Hz and then queries again.  The arm holds still while the next chunk is computed.

\parahead{Trial protocol.}
Before each trial, the arm returns to a fixed home pose with the gripper open, and the operator places the objects.  Their positions vary around the demonstration layouts.  A trial ends at success or after 60\,s.  We run 10 trials per task and policy, and Tab.~\ref{table:real-robot-tasks} defines success and score.

\end{document}